\documentclass[11pt]{article}

\usepackage[final]{acl}

\usepackage{times}
\usepackage{latexsym}

\usepackage[T1]{fontenc}

\usepackage[utf8]{inputenc}

\usepackage{microtype}
\usepackage{booktabs}

\usepackage[most]{tcolorbox}
\usepackage{listings}
\usepackage{floatrow}
\usepackage[table]{xcolor}
\usepackage[dvipsnames]{xcolor} 
\usepackage{pifont} 
\usepackage{caption}
\DeclareCaptionFont{tightcaption}{\fontsize{9pt}{8.8pt}\selectfont}
\DeclareFloatSeparators{mysep}{\hskip 0.5em}

\usepackage{amsmath}
\usepackage{amssymb}
\usepackage{mathtools}
\usepackage{amsthm}

\usepackage{inconsolata}

\usepackage{graphicx}

\title{\includegraphics[width=0.765cm]{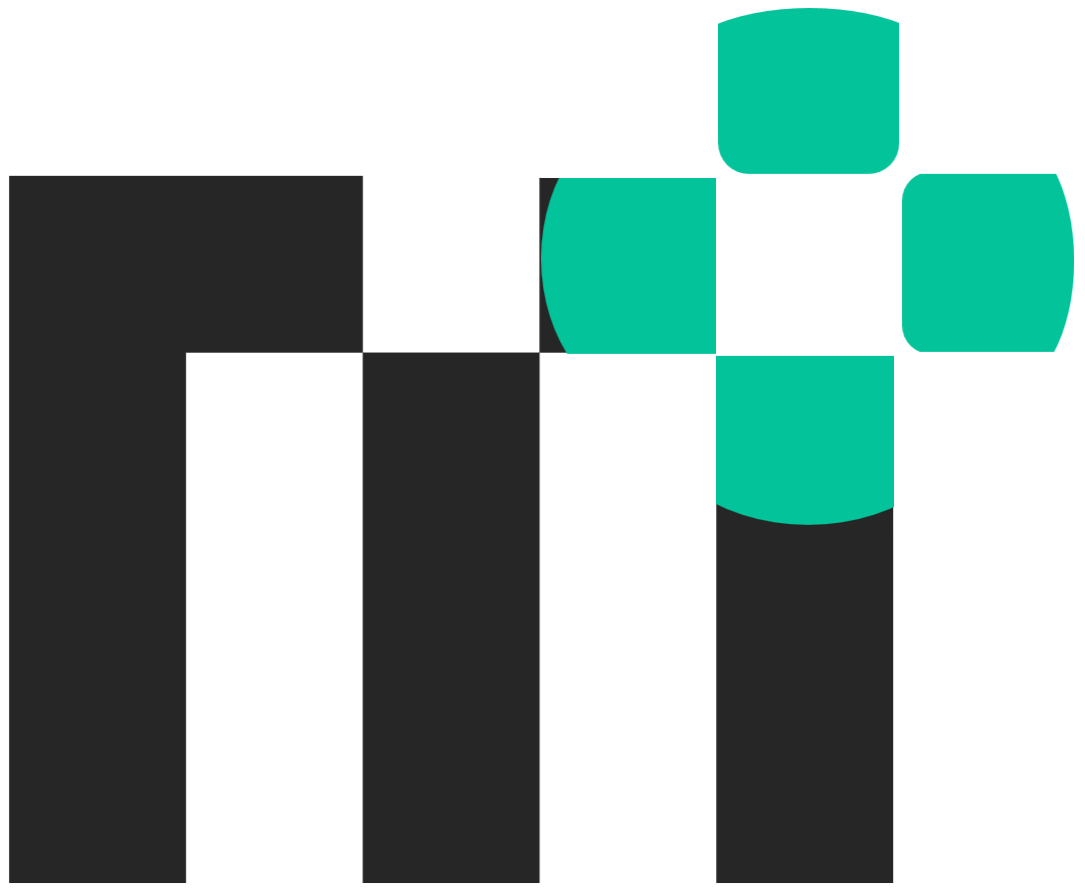} \hspace{0.2mm}
  MediX-R1:  Open Ended Medical Reinforcement Learning}

\author{Sahal Shaji Mullappilly$^1$\thanks{Equal Contribution}, ~Mohammed Irfan$^1$\footnotemark[1], ~Jaseel Muhammad$^1$, \textbf{~Omair Mohamed$^3$}, \\\textbf{~Mohamed Zidan$^4$, ~Fahad Khan$^{1,2}$}\textbf{~Salman Khan$^1$, ~Rao Anwer$^1$, ~Hisham Cholakkal$^1$} \vspace{1mm} \\ 
$^1$Mohamed Bin Zayed University of Artificial Intelligence (MBZUAI), $^2$Linköping University\\ $^3$Cape Fear Valley Medical Center, $^4$East Point College of Medical Sciences \& Research Centre
}

\begin{document}

\newsavebox{\herobox}
\savebox{\herobox}{\includegraphics[width=\textwidth]{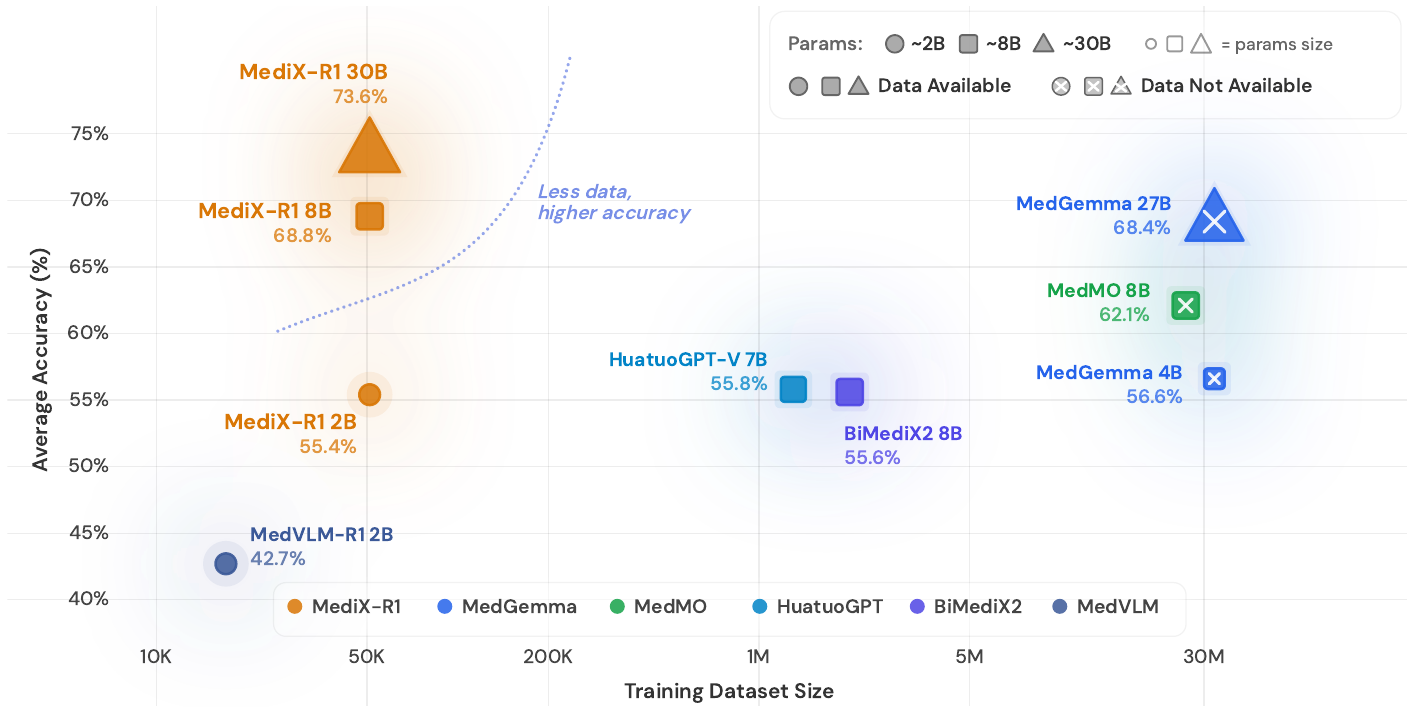}}

\makeatletter
\begingroup
\def\thefootnote{\fnsymbol{footnote}}
\twocolumn[
  \@maketitle
  \begin{center}
    \usebox{\herobox}
  \captionof{figure}{\textbf{Average accuracy across multimodal medical benchmarks vs.\ training dataset size for recent medical VLMs.} Colors denote model families; marker shape/size indicates parameter scale $\sim$(2B, 8B, 30B).  × indicates the absence of public training data.  MediX-R1 8B (68.8\%) surpasses MedGemma 27B (68.4\%) while using significantly less training data, and MediX-R1 30B achieves the highest overall accuracy (73.6\%).
  }
  \label{fig:scatter_hero}
  \end{center}
]
\@thanks
\endgroup
\setcounter{footnote}{0}
\makeatother

\begin{abstract}
In this work, we introduce MediX-R1, an open-ended Reinforcement Learning (RL) framework for medical vision language models (VLMs) that enables free-form answers beyond multiple-choice formats. MediX-R1 fine-tunes a baseline vision-language backbone with Group Based RL and a composite reward tailored for medical reasoning: an LLM-based accuracy reward that judges semantic correctness with a strict YES/NO decision, a medical embedding-based semantic reward to capture paraphrases and terminology variants, and lightweight format and modality rewards that enforce interpretable reasoning and modality recognition. This multi-signal design provides stable, informative feedback for open-ended outputs where traditional verifiable or MCQ-only rewards fall short. To evaluate performance, we employ a unified evaluation framework for both text-only and multimodal (image+text)  tasks that uses a Reference-based LLM-as-judge in place of brittle string-overlap metrics, capturing semantic correctness, clinical accuracy, and contextual alignment. Our results show that open-ended RL with comprehensive reward signals can effectively strengthen medical reasoning in multimodal models. Despite using only $\sim51$K instruction examples, MediX-R1 achieves strong performance across standard medical LLM (text-only) and VLM (image + text) benchmarks, outperforming existing open-weight medical VLMs, with large gains particularly on open-ended clinical tasks. 
These improvements remain consistent across multiple RL algorithms and VLM backbones and are further supported by blinded evaluations with medical experts.
Our \textit{trained models}, \textit{curated datasets} and \textit{source code} are available at \href{https://medix.cvmbzuai.com}{\textbf{MediX-R1}}.
\end{abstract}

\begin{figure*}[t!]
  \centering
    \includegraphics[width=\linewidth]{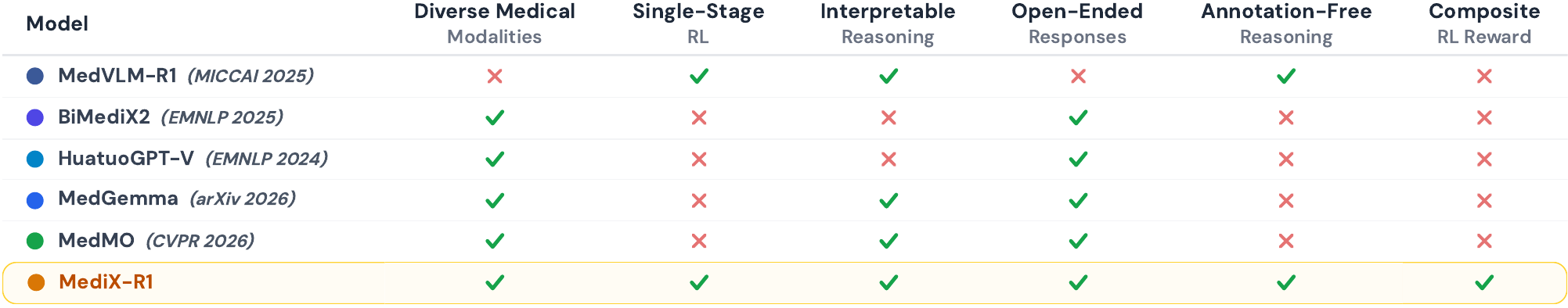}
\caption{\textbf{Differences with existing Medical VLMs: } 
MediX-R1 integrates diverse modalities, interpretable reasoning, and composite RL rewards, enabling practical clinical use.}
\label{fig:model-capability}
\end{figure*}

\section{Introduction}
Large medical language and vision-language models are increasingly deployed for clinical question answering, triage support, report drafting, and education \citep{chen2024huatuogpto1medicalcomplexreasoning, sellergren2025medgemma, pieri-etal-2024-bimedix}. Many of these tasks are inherently open-ended: clinicians expect concise but free-form answers that can flexibly incorporate context, uncertainty, and multimodal evidence. However, most training and evaluation pipelines remain tailored to Multiple Choice Questions (MCQ) or string-matching regimes, which (i) under-reward valid clinical paraphrases, (ii) fail to measure reasoning quality or modality recognition, and (iii) do not provide reliable signals for reinforcement learning (RL) in open-ended settings. As a result, models trained only with supervised objectives or MCQ-style rewards often struggle to produce faithful, interpretable, and robust clinical responses across diverse medical tasks.

RL has improved reasoning in domains with verifiable rewards (e.g., math and code) as shown by DeepSeek models \citep{shao2024deepseekmath, guo2025deepseek}, but medical tasks rarely admit executable checks. Binary exact match is too brittle for clinical phrasing; BLEU/ROUGE can misjudge semantically correct answers; and free-form VLM outputs complicate visual inference. Moreover, using a single reward signal can induce instability or reward hacking, especially when the signal is noisy or overly permissive. Hence, it is desirable to have a systematic approach for training medical VLMs with open-ended RL that integrates semantic correctness with structural and modality constraints, while remaining data- and compute-efficient.

In this work, we propose MediX-R1, an open-ended medical RL framework that fine-tunes a baseline multimodal backbone using a group-based RL algorithm with a composite reward tailored for clinical reasoning and applicable across different RL optimization strategies (e.g., GRPO, GSPO, and DAPO).
The proposed  composite reward combines: \textit{(1)} an LLM-based accuracy reward that enforces a strict YES/NO decision on semantic correctness, \textit{(2)} a medical embedding-based semantic reward that captures paraphrases and terminology variants, \textit{(3)} a lightweight format reward that elicits interpretable reasoning traces, and \textit{(4)} a modality recognition reward that discourages cross-modality hallucinations by requiring explicit modality tags. This multi-signal objective stabilizes optimization and gives informative feedback where traditional verifiable or MCQ-only rewards fall short, enabling single-stage, open-ended RL on clinical tasks.

\begin{figure*}[t!]
  \centering
    \includegraphics[width=\linewidth]{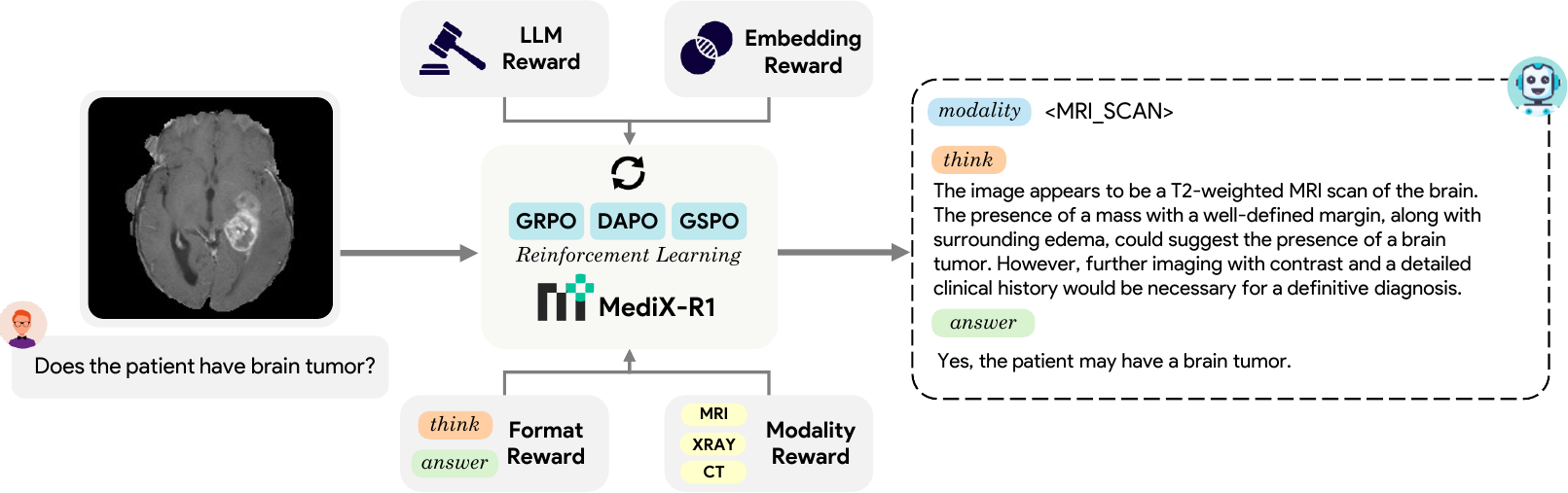}
\caption{\textbf{MediX-R1: Overall Architecture} The MediX-R1 reinforcement learning framework for open-ended medical reasoning. An input of a medical image and a natural language question is processed by MediX-R1. The model's policy is trained using Group Based RL, which leverages a multi-faceted reward signal. This reward is composed of: a) an LLM-based reward for evaluating the quality and accuracy of the output; b) an embedding-based reward to ensure semantic alignment; c) a format reward to enforce the desired output structure ({\footnotesize\texttt{<think>}} and {\footnotesize\texttt{<answer>}} blocks); and d) a modality reward to ensure the response is grounded in the specified imaging modality. This reward-guided approach encourages the model to generate accurate and interpretable reasoning paths.}
\end{figure*}

\noindent\textbf{Differences with existing Medical VLMs:}
Figure~\ref{fig:model-capability} contrasts MediX-R1 with strong open models across key clinical capabilities. First, on \emph{Diverse Medical Modalities}, MediX-R1 supports diverse medical modalities including X-Ray, CT, MRI, Microscopy/Histopathology, Ultrasound, Fluoroscopy, Endoscopy, Angiography, Mammography, Clinical Photography, SPECT (Single Photon Emission Computed Tomography), OCT (Optical Coherence Tomography), and Fundus imaging, whereas MedVLM-R1~\citep{pan2025medvlm} is limited to radiology images. Models like MedGemma~\citep{sellergren2026medgemma15technicalreport}, HuatuoGPT-Vision~\citep{chen2024huatuogptvisioninjectingmedicalvisual}, MedMO~\citep{deria2026medmogroundingunderstandingmultimodal}, and BiMediX2~\citep{mullappilly2024bimedix2biomedicalexpertlmm} provide coverage on clinical modalities but they require extensive multi-stage training. On \emph{Single-Stage RL}, most baselines rely on multi-stage pipelines (pretraining $\rightarrow$ SFT $\rightarrow$ RL), whereas MediX-R1 is trained end-to-end with a single RL stage using our composite reward (Sec.~\ref{subsec:rewards}). This simplifies training and, importantly, enables \emph{open-ended} RL directly (unlike MedVLM-R1), because the Reference-based LLM-as-judge accuracy signal and medical embeddings provide reliable feedback beyond MCQ exact match. The composite design (format + LLM judge + embeddings + modality recognition) stabilizes optimization and reduces reward hacking (Fig.~\ref{fig:reward_design}), translating into the best average performance in Table~\ref{tab:eval-benchmarks}. For \emph{Interpretable Reasoning}, MediX-R1 generates explicit reasoning traces enclosed in {\small\texttt{<think>}...\texttt{</think>}}, enforced by the format reward, making the decision path auditable. Several  baselines  do not reliably produce structured clinical rationales. While multiple models support \emph{Open-Ended Responses}, MediX-R1 is explicitly optimized for free-form clinical answering with modality recognition, which reduces cross-modality hallucinations and improves VLM robustness. MediX-R1 achieves \emph{Annotation-Free Reasoning}: it does not require human-curated rationales or verified chain-of-thought. The RL rewards operate on the final answer only (via Reference-based LLM judge and embeddings), significantly lowering data curation cost while still encouraging interpretable reasoning. Collectively, these properties contribute to MediX-R1’s strong performance across multiple  medical benchmarks and enhance its practical effectiveness for real-world deployment.

For medically relevant performance evaluation, instead of relying on brittle metrics such string-overlap and binary-matching, we employ a \textbf{\textit{reference-based LLM-as-judge}} framework that leverages an LLM to assess the semantic correctness and contextual alignment of \textit{model predictions against corresponding question-specific ground-truth answers} through a unified three-stage evaluation process. The evaluations span tasks ranging from short-form QA to long-form report generation, thereby reducing the mismatch between benchmark evaluation and clinical utility.
Despite using only $\sim$51K instruction samples (Fig~\ref{fig:scatter_hero}, \S\ref{app:training-data}), MediX-R1 achieves strong results across diverse medical benchmarks. Our experiments show that the proposed composite rewards not only improve benchmark accuracy,  but also mitigate reward hacking and reduce volatility, yielding stable training and interpretable outputs. Compared to open-weight  medical models listed in Fig~\ref{fig:model-capability}, MediX-R1 combines broad modality coverage 
with single-stage RL and structured reasoning. In this work we introduce a domain-specific training and evaluation framework that extends existing group-based RL algorithms to open-ended medical vision-language tasks using a composite reward designed for clinical accuracy, output structure, and modality recognition.

\begin{figure*}[ht!]
\captionsetup{font+=tightcaption}
\floatbox[{\capbeside\thisfloatsetup{capbesideposition={right,top},capbesidewidth=0.26\textwidth, capbesidesep=mysep}}]{figure}[\FBwidth]
{%
  \caption{\textbf{Evaluation Framework:} Our three-stage evaluation pipeline: (1) Generation via vLLM inference on the model under test, (2) Evaluation using Reference-based LLM-as-judge with BASE and MIMIC templates, and (3) Scoring through aggregation of judgment outputs. The framework supports both binary decisions for QA/MCQ tasks and rubric-based scoring for long-form reports, ensuring robust evaluation across diverse medical benchmarks.}%
  \label{fig:eval-framework}
} 
{%
  \includegraphics[width=0.72\textwidth]{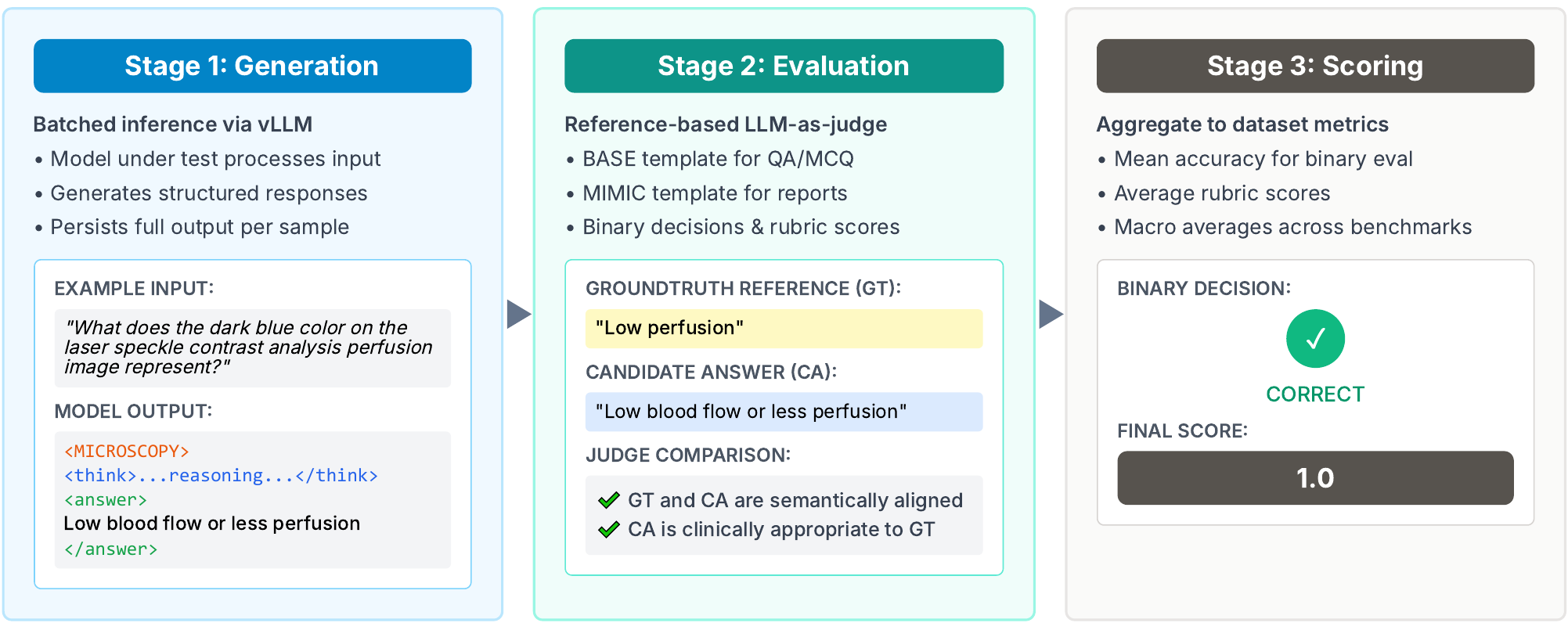}%
}
\end{figure*}

\noindent\textbf{Contributions:}  \textbf{(i)} We introduce \textit{open-ended medical reinforcement learning} by extending Group based RL with tailored rewards for clinical accuracy. \textbf{(ii)} We design a \textit{composite reward} with LLM-based accuracy and medical semantic signals that  enables open-ended responses with RL in the medical domain and stabilizes training. \textbf{(iii)} We employ a three-stage \textit{Reference-based LLM-as-judge evaluation framework} that unifies benchmarking for both LLM (text-only) and VLM (image+text) tasks in the medical setting. \textbf{(iv)} MediX-R1 achieves excellent LLM and VLM results with a \textit{single-stage RL recipe using $\sim$51K instructions}, validated through both Reference-based LLM-as-judge and human expert evaluations. \textbf{(v)}  Finally, we show the effectiveness of the proposed composite reward across different  Group based RL algorithms, achieving consistent performance gains with GRPO~\citep{shao2024deepseekmath}, DAPO~\citep{yu2025dapoopensourcellmreinforcement} and GSPO~\citep{zheng2025groupsequencepolicyoptimization}. Moreover, we have conducted experiments on different baseline VLMs, including Qwen2.5-VL, Qwen3-VL~\citep{bai2025qwen3vltechnicalreport}, and SmolVLM2~\citep{marafioti2025smolvlmredefiningsmallefficient}, and achieved consistent performance gains across these backbones.

\section{Open Ended Medical RL}

MediX-R1 fine-tunes a baseline multimodal backbone for open-ended medical reasoning using Group Based RL. Given an image $I$ and question $q$, the vision encoder produces visual tokens that are fused with text tokens and fed to the LLM policy $\pi_\theta$. The model generates structured outputs of the form: {\fontsize{8pt}{9pt}\selectfont\texttt{[modality tag]<think>free-form clinical reasoning</think><answer>final concise answer</answer>}}

\subsection{\fontsize{10.5pt}{9pt}\selectfont{Group-based RL with Composite Rewards}}
\textbf{\underline{Setup:}}
Given an input $\mathbf{v}$ (image $I$ + question $q$) drawn from $P(\mathbf{V})$, we sample a group of $G$ candidate completions $\{o_i\}_{i=1}^{G}$ from the frozen behavior policy $\pi_{\theta_{\text{old}}}(\cdot\mid \mathbf{v})$. Each completion receives a scalar reward $r_i$ computed by our \emph{composite reward} (Sec.~\ref{subsec:rewards}). We then compute a standardized \emph{group-relative} advantage:
{\footnotesize
\[
A_i \;=\; \frac{r_i - \mathrm{mean}(\{r_j\}_{j=1}^G)}{\mathrm{std}(\{r_j\}_{j=1}^G)}.
\]
}

\noindent This removes the need for a learned value function while preserving a stable relative learning signal within group.

\noindent\textbf{\underline{GRPO objective:}}
GRPO~\citep{shao2024deepseekmath} updates $\pi_\theta$ using PPO-style clipping on an importance ratio and a KL regularizer to a fixed ref policy $\pi_{\text{ref}}$:
{\scriptsize
\begin{equation}
\begin{split}
\mathcal{J}_{\text{GRPO}}(\theta)
=\mathbb{E}_{\mathbf{v},\,\{o_i\}}\Bigg[
\frac{1}{G}\sum_{i=1}^G 
\min\!\Big(\rho_i(\theta)A_i, \\\;\mathrm{clip}(\rho_i(\theta),\,1-\epsilon,\,1+\epsilon)A_i\Big)
-\beta\,\mathbb{D}_{\mathrm{KL}}\!\left(\pi_{\theta}\,\|\,\pi_{\text{ref}}\right)
\Bigg]
\end{split}
\label{eq:GRPO}
\end{equation}
}

\noindent where $\rho_i(\theta)=\frac{\pi_\theta(o_i\mid \mathbf{v})}{\pi_{\theta_{\text{old}}}(o_i\mid \mathbf{v})}$, and $\epsilon,\beta\ge 0$ control clipping and regularization strength.

\noindent\textbf{\underline{Composite reward with different optimizers:}}
In addition to GRPO, we run the \emph{same} composite reward with two recent Group based RL family optimizers, DAPO~\citep{yu2025dapoopensourcellmreinforcement} and GSPO~\citep{zheng2025groupsequencepolicyoptimization} and report their comparison in Table~\ref{tab:rl-method-ablation}. 

\noindent\textbf{\underline{DAPO (efficiency-focused refinements):}}
DAPO keeps the GRPO/PPO clipped structure but improves token efficiency (as summarized in \citet{yu2025dapoopensourcellmreinforcement}): it (i) uses \emph{asymmetric clipping} (“Clip-Higher”) with a larger upper bound to avoid prematurely zeroing gradients for rare-but-good tokens, and (ii) averages loss over \emph{all generated tokens} rather than per-sample averaging (reducing gradient dilution for long responses). Compact form:
{\scriptsize
\begin{equation}
\begin{split}
\mathcal{J}_{\text{DAPO}}(\theta)
=\mathbb{E}_{\mathbf{v},\,\{o_i\}}
\Bigg[
\frac{1}{\sum_{i=1}^G |o_i|}\sum_{i=1}^G\sum_{t=1}^{|o_i|} 
\min\!\Big(r_{i,t}(\theta)A_i, \\\;\mathrm{clip}(r_{i,t}(\theta),\,1-\epsilon_{\text{low}},\,1+\epsilon_{\text{high}})A_i\Big)
\Bigg]
\end{split}
\label{eq:DAPO}
\end{equation}
}
\noindent where $r_{i,t}(\theta)=\frac{\pi_\theta(o_{i,t}\mid \mathbf{v},o_{i,<t})}{\pi_{\theta_{\text{old}}}(o_{i,t}\mid \mathbf{v},o_{i,<t})}$.
DAPO further encourages rollouts via \emph{dynamic sampling} (avoiding degenerate groups with identical rewards) and can apply \emph{overlong reward shaping} to softly penalize excessively long outputs.

\noindent\textbf{\underline{GSPO (sequence-level ratio for stability):}}
GSPO addresses variance from token-wise importance ratios by replacing $r_{i,t}(\theta)$ with a \emph{length-normalized sequence-level ratio} shared across all tokens in a completion.
It then uses the same clipped objective form but with $s_i(\theta)$ in place of token-wise ratios:
{\scriptsize
\begin{equation}
\begin{split}
\mathcal{J}_{\text{GSPO}}(\theta)
=\mathbb{E}_{\mathbf{v},\,\{o_i\}}\Bigg[
\frac{1}{G}\sum_{i=1}^G
\min\!\Big(s_i(\theta)A_i, \\\;\mathrm{clip}(s_i(\theta),\,1-\epsilon,\,1+\epsilon)A_i\Big)
\Bigg]
\end{split}
\label{eq:GSPO}
\end{equation}
}

\begin{table*}[t!]
\centering
\resizebox{\textwidth}{!}{%
\begin{tabular}{@{}lccccccccc@{}}
\toprule
\textbf{Benchmarks} &
\multicolumn{1}{c}{\textbf{\shortstack{MedVLM\\R1 2B}}} &
\multicolumn{1}{c}{\textbf{\shortstack{BiMediX2\\8B}}} &
\multicolumn{1}{c}{\textbf{\shortstack{Huatuo\\GPT-V 7B}}} &
\multicolumn{1}{c}{\textbf{\shortstack{MedGemma\\1.5 4B}}} &
\multicolumn{1}{c}{\textbf{\shortstack{MedGemma\\27B}}} &
\multicolumn{1}{c}{\textbf{\shortstack{MedMO\\8B}}} &
\multicolumn{1}{c}{\textbf{\shortstack{MediX-R1\\2B}}} &
\multicolumn{1}{c}{\textbf{\shortstack{MediX-R1\\8B}}} &
\multicolumn{1}{c}{\textbf{\shortstack{MediX-R1\\30B}}} \\
\midrule
MMLU-Clinical    & 0.540 & 0.732 & 0.721 & 0.706 & \underline{0.879} & 0.864 & 0.660 & 0.845 & \textbf{0.894} \\
MMLU-Bio         & 0.549 & 0.792 & 0.708 & 0.708 & \underline{0.972} & 0.951 & 0.806 & 0.951 & \textbf{0.993} \\
MMLU-Med         & 0.451 & 0.694 & 0.653 & 0.578 & 0.866 & 0.827 & 0.699 & \underline{0.879} & \textbf{0.890} \\
MMLU-Genetics    & 0.560 & 0.790 & 0.710 & 0.760 & \underline{0.940} & 0.900 & 0.680 & 0.900 & \textbf{0.980} \\
MMLU-ProfMed     & 0.500 & 0.695 & 0.625 & 0.768 & \underline{0.912} & 0.890 & 0.628 & 0.868 & \textbf{0.974} \\
MMLU-Anat        & 0.519 & 0.659 & 0.600 & 0.519 & \underline{0.793} & 0.785 & 0.563 & 0.763 & \textbf{0.874} \\
MedMCQA          & 0.408 & 0.572 & 0.511 & 0.538 & \underline{0.727} & 0.662 & 0.492 & 0.683 & \textbf{0.781} \\
MedQA            & 0.400 & 0.583 & 0.534 & 0.690 & \underline{0.866} & 0.848 & 0.497 & 0.796 & \textbf{0.929} \\
USMLE-SA         & 0.378 & 0.591 & 0.538 & 0.726 & \underline{0.895} & 0.742 & 0.505 & 0.822 & \textbf{0.951} \\
PubMedQA         & 0.520 & 0.520 & \underline{0.542} & 0.460 & 0.414 & \textbf{0.586} & 0.472 & 0.482 & 0.490 \\
MIMIC-CXR-Sum    & 0.704 & 0.672 & 0.707 & 0.699 & \textbf{0.767} & 0.709 & 0.739 & 0.746 & \underline{0.765} \\
\midrule
SLAKE-VQA        & 0.434 & 0.468 & 0.545 & 0.565 & 0.634 & 0.479 & 0.654 & \textbf{0.703} & \underline{0.683} \\
RadVQA           & 0.404 & 0.530 & \underline{0.614} & 0.572 & 0.585 & 0.419 & 0.539 & 0.596 & \textbf{0.625} \\
PathVQA          & 0.239 & 0.323 & 0.374 & 0.332 & 0.322 & 0.272 & 0.428 & \textbf{0.455} & \underline{0.445} \\
PMC-VQA          & 0.398 & 0.482 & 0.532 & 0.474 & 0.478 & 0.360 & 0.491 & \underline{0.554} & \textbf{0.571} \\
PMC-VQA-Hard     & 0.020 & 0.229 & 0.261 & 0.258 & \textbf{0.354} & 0.177 & 0.284 & \underline{0.317} & 0.307 \\
MIMIC-CXR-Gen    & 0.240 & 0.124 & 0.316 & 0.273 & 0.224 & 0.084 & 0.280 & \underline{0.328} & \textbf{0.350} \\
\midrule
\textbf{AVG}     & 0.427 & 0.556 & 0.558 & 0.566 & 0.684 & 0.621 & 0.554 & \underline{0.688} & \textbf{0.736} \\
\bottomrule
\end{tabular}%
}
\caption{\textbf{Evaluation Benchmark:} The top section lists LLM (text-only) tasks and the bottom section lists VLM (image+text) tasks. Our three-stage evaluation setting evaluates both tasks in a unified framework. MediX-R1 achieves the highest average score across this diverse suite, exhibiting state-of-the-art performance among open models. Best and second best results are bold and underlined. Task specific results shown in Table~\ref{tab:task-specific-results}.}
\label{tab:eval-benchmarks}
\end{table*}

\noindent Because all tokens within a sequence share the same ratio, GSPO reduces intra-sequence variance and tends to be more stable in settings where token-level ratios are noisy (e.g., long outputs and dynamic architectures).

\subsection{Reward Design}
\label{subsec:rewards}
We define a composite reward:
{\small
\[
r = w_{\text{fmt}}\,R_{\text{format}} + w_{\text{llm}}\,R_{\text{llm}}+ w_{\text{emb}}\,R_{\text{embed}} + w_{\text{mod}}\,R_{\text{modality}}
\]
}

\noindent with coefficients chosen via a simple, staged ablation procedure. In all experiments we keep a small, fixed weight on the format reward to enforce parseable outputs, and distribute the remaining weight across semantic correctness (LLM judge / embeddings) and modality grounding. The coefficient selection procedure is described in Sec.~\ref{subsec:reward-hacking}, and full training config in \S\ref{app:training-config}.

\noindent\textit{Why this enables open-ended medical RL :} Unlike prior RL setups that are limited to verifiable signals or MCQ-style accuracy (e.g., exact match, executable or rule-based graders), our LLM-based accuracy reward $R_{\text{llm}}$ and embedding-based semantic reward $R_{\text{embed}}$ provide reliable feedback for free-form, clinically accurate answers. The Reference-based LLM-as-judge converts semantic correctness into a robust YES/NO decision under paraphrase and clinical phrasing, while medical-domain embeddings supply a complementary content-alignment signal. This dual signal makes Reinforcement Learning viable for open-ended medical reasoning; the format ($R_{\text{format}}$) and modality ($R_{\text{modality}}$) rewards act as structural regularizers, but $R_{\text{llm}}$ and $R_{\text{embed}}$ are the primary drivers of open-ended RL in MediX-R1. Our composite reward encourages structured and grounded outputs, but does not directly optimize or guarantee faithful intermediate reasoning.

\noindent\textbf{{LLM-Based Accuracy Reward ($R_{\text{llm}}$) :}}~\mbox{We parse} the model output’s final answer between {\footnotesize\texttt{<answer>...</answer>}} and compare it to the reference answer using a compact Reference-based LLM-as-judge prompt that forces a strict YES/NO decision. Concretely, a local vLLM ~\citep{kwon2023efficient} endpoint (Qwen3-4B) returns YES if the candidate semantically answers the reference, and NO otherwise; we map YES$\mapsto 1$, NO$\mapsto 0$. This captures correctness and robustness to paraphrasing while keeping the signal discrete.

\noindent\textbf{Embedding-Based Reward ($R_{\text{embed}}$) :}
To further encourage semantic alignment, we compute cosine similarity between the predicted answer and the reference using a medical embedding model MedEmbed-large \citep{balachandran2024medembed}. We convert it to a binary reward via a threshold (default 0.8): $R_{\text{embed}}{=}1[\cos(\mathbf{e}_{\text{pred}},\mathbf{e}_{\text{ref}})\ge\tau]$. This complements the LLM judge and helps capture terminological variants. Our threshold ablation in Table~\ref{tab:embedding-threshold-ablation} shows that $\tau=0.8$ provides the best balance between precision and recall, avoiding false-positive rewards at lower thresholds while retaining valid semantic paraphrases missed at higher thresholds.

\noindent\textbf{Format Reward ($R_{\text{format}}$) :}
We enforce structured outputs by matching the regex for the exact pattern
{\footnotesize\texttt{<think>...</think>}} {\footnotesize\texttt{<answer>...</answer>}}
after normalizing stray whitespace around brackets. Outputs that match receive $1$, else $0$. This stabilizes training and improves interpretability of the medical reasoning. The format reward enforces the presence and structure of models output, but no reward directly evaluates the thinking process and its clinical correctness, or correspondence to the model’s internal decision process.

\noindent\textbf{Modality recognition Reward ($R_{\text{modality}}$) :}
We encourage explicit grounding to the imaging modality by requiring the model to emit the predicted modality tag before the {\small\texttt{</think>}} block (case-insensitive). We compare it to the reference modality tag and assign $1$ on match, $0$ otherwise. This reduces cross-modality hallucinations (e.g., describing CT findings on an X-ray).

\section{Evaluation Framework}

Our evaluation pipeline has three stages: Generation, Evaluation, and Scoring. We evaluate across both text-only (LLM) and image+text (VLM) tasks covering QA, MCQ, and long-form report tasks.

\noindent\textbf{Generation.} We run batched inference via vLLM on the model under test and persist the full response per sample. For models that emit structured reasoning, we retain the entire output but, for scoring, discard internal chains-of-thought by stripping content up to and including the closing {\small\texttt{</think>}} tag, evaluating only the final answer block.

\noindent\textbf{Evaluation:} We employ a separate Reference-based LLM-as-judge, Qwen3-14B \citep{yang2025qwen3technicalreport}, served with vLLM for throughput and stability on modest GPUs. Two prompt families are used: a BASE template (\S\ref{lst:eval-base}) for open-ended, one-word, and MCQ-style questions that yields a binary decision, and a MIMIC template (\S\ref{lst:eval-report}) for long-form report generation that scores along clinical criteria. For example, on a visual question answering item asking “\textit{What does the dark blue color on the laser speckle contrast analysis perfusion image represent?}” with ground truth “\textit{Low perfusion}” a model response that includes hidden reasoning and the final answer “\textit{Low blood flow or less perfusion}” is judged correct and assigned a score of 1. The judge compares predicted answers against references, accounting for paraphrase and clinically equivalent phrasing. 
\textbf{Scoring:} We aggregate judgment outputs to dataset-level metrics. For binary evaluations, we report mean accuracy over samples. For long-form, we average the scalar rubric scores across samples, optionally normalizing for comparability. We also compute macro averages across benchmarks.
\textbf{Why Reference-based LLM-as-judge  (via vLLM):} Traditional string-overlap metrics (BLEU, ROUGE, F1) for reference (ground truth)  comparison often under-reward correct, clinically appropriate paraphrases and cannot assess justification quality or contextual alignment. A Reference-based LLM judge captures semantic correctness, clinical accuracy, and adherence to task-specific criteria through carefully designed prompts, while vLLM serving ensures consistent, fast, and reproducible evaluations.

\begin{table}[t!]
\centering
\footnotesize
\resizebox{\textwidth}{!}{%
\begin{tabular}{@{}l c@{}}
\toprule
\textbf{Model} & \textbf{MMMU Medical Validation} \\
\midrule
MedVLM-R1 2B \citep{pan2025medvlm}                                   & 39.33 \\
MedGemma 1.5 4B \citep{sellergren2026medgemma15technicalreport}      & 50.00 \\
HuatuoGPT-V 7B \citep{chen2024huatuogptvisioninjectingmedicalvisual} & 47.33 \\
BiMediX2 8B \citep{mullappilly2024bimedix2biomedicalexpertlmm}       & 39.33 \\
Qwen3-VL 8B \citep{bai2025qwen3vltechnicalreport}                    & 62.66 \\
MedMO 8B \citep{deria2026medmogroundingunderstandingmultimodal}     & 62.66 \\
Qwen3-VL 30B \citep{bai2025qwen3vltechnicalreport}                   & 68.66 \\
MedGemma 27B \citep{sellergren2025medgemma}                & 56.66 \\
Lingshu 32B \citep{lasateam2025lingshugeneralistfoundationmodel}     & 62.66 \\
MediX-R1 2B                                                 & 44.66 \\
MediX-R1 8B                                                 & 64.00 \\
MediX-R1 30B                                                & 75.33 \\
\bottomrule
\end{tabular}
}
\caption{\textbf{MMMU Medical Val results:} Accuracy on the Health \& Medical validation subset of MMMU.}
\label{tab:mmmu-med-val}
\end{table}

\section{Experiments and Results}
\subsection{State-of-the-art Comparisons}
We evaluate MediX-R1 on a comprehensive suite of medical language and vision-language benchmarks, covering both text-only (LLM) and image+text (VLM) tasks. The evaluation includes standard medical QA, multiple-choice, and open-ended report generation, as well as visual question answering and clinical image interpretation. The datasets used for evaluation are listed in \S\ref{app:training-data}.

For each dataset, we follow the evaluation as described in the previous section, using Reference-based LLM-as-judge scoring for both short-form and long-form responses. Table~\ref{tab:eval-benchmarks} summarizes performance on our unified benchmark suite across several open-weight medical models, including MedVLM-R1 \textit{(2B)}, BiMediX2 \textit{(8B)}, HuatuoGPT-V \textit{(7B)}, MedGemma \textit{(4B/27B)}, MedMO \textit{(8B)} \citep{deria2026medmogroundingunderstandingmultimodal} and MediX-R1 \textit{(2B/8B/30B)}.

MediX-R1 achieves the highest average score across all benchmarks, outperforming prior models on both language and vision-language tasks. Notably, it shows strong gains on open-ended and clinically complex tasks such as report generation, as well as robust performance on standard QA, VQA and crowdsourced real-world benchmarks. These results highlight the effectiveness of our open-ended RL training and composite reward design, which enable MediX-R1 to generate accurate, semantically aligned, and clinically auditable responses beyond the capabilities of models trained only with supervised or MCQ objectives. Since the evaluation suite covers heterogeneous capabilities, we additionally report task-specific results for text-only QA/MCQ, image-text VQA/MCQ, report generation, and report summarization in Appendix~\ref{app:task-specific-results}. Models are grouped into $\leq$2B, $\leq$8B, and $\leq$30B parameter classes to enable size-matched comparisons, showing that MediX-R1 achieves consistent gains across most task categories and model scales rather than benefiting from a single task type.

We additionally report results on the Massive Multi-discipline Multimodal Understanding and Reasoning (MMMU) benchmark \citep{yue2024mmmumassivemultidisciplinemultimodal}. We select the Health \& Medical validation subset (MMMU-Med-Val), covering Basic Medical Science, Clinical Medicine, Diagnostics  and Laboratory Medicine, Pharmacy, and Public Health. Results are shown in Table~\ref{tab:mmmu-med-val}. 

\subsection{Ablation Experiments}

\noindent\textbf{Composite Reward across RL algorithms}: Our RL-method ablation in Table~\ref{tab:rl-method-ablation} shows that the proposed composite-reward training transfers across different RL frameworks, consistently improving over the baseline across both LLM and VLM tasks. Using Qwen2.5-VL as the baseline, DAPO achieves the best overall average (0.610), compared with GRPO (0.597), GSPO (0.600), and the baseline model (0.570). We observe a similar trend with the stronger Qwen3-VL 8B baseline, where GRPO, GSPO, and DAPO improve the overall average from 0.666 (baseline) to 0.679, 0.681, and 0.688, respectively. DAPO remains the strongest method across both backbones, showing the consistent effectiveness of the composite reward across RL algorithms and model backbones.

\noindent\textbf{Reward Design Ablation: } 
Table~\ref{tab:reward-ablation} compares reward variants that differ in which non-format signals are active (all settings include the same $R_{\text{format}}$ and the default string match reward). The Default reward baseline that only uses basic string matching against the ground truth is brittle to paraphrases and clinical synonymy. Using only the embedding reward underperforms on text-only evaluations (0.640) and provides gains on VLM (0.409), suggesting that thresholded cosine similarity alone lacks discriminative power for nuanced clinical reasoning. Using only the LLM judge improves text-only accuracy (0.666) but does not help VLM (0.400), indicating that the judge alone is insufficient to enforce modality grounding. All reward design models are compared with checkpoints before reward hacking.

Combining LLM + embedding increases robustness to paraphrase and terminology variants, improving text-only scores (0.686) and yielding a small VLM lift (0.410), which raises the overall average to 0.589. Finally, the full MediX-R1 composite (LLM accuracy + embedding semantics + modality recognition, with shared format control) produces the strongest image+text performance (0.431) while matching the best text-only result (0.687), achieving the best overall average (0.597). Together with Fig.~\ref{fig:reward_design}, which shows reduced volatility and fewer signs of reward hacking, these results suggest that the composite reward both improves aggregate performance and stabilizes optimization.

Key takeaways: \textit{(i)} The LLM judge is the strongest single signal for text correctness, and embeddings complement it by reducing false negatives from paraphrases.\textit{(ii)} Default string matching degrades substantially on open-ended multimodal evaluations.\textit{(iii)} Modality recognition is important for VLM tasks and drives gains in image+text tasks; the full composite delivers the best overall results.

\noindent\textbf{Performance across VLM backbones}: We observe consistent improvements across VLM backbones after training these models in our MediX-R1 framework with our composite rewards, as shown in Table~\ref{tab:baseline-comparison}. These results show that MediX-R1 enhances open-ended medical reasoning ability across backbone models and model scales without any additional parameters.

\noindent\textbf{Judge Robustness and Evaluator sensitivity.}
To ensure robustness, we perform controlled evaluations using deterministic generation settings {\footnotesize(\texttt{temperature=0}, \texttt{top\_p=1})} and report averages over three runs, observing only $\pm 0.002$ variation. For additional validation, we replaced Qwen3-14B with GPT-5.1 and GPT-5 mini as evaluators, which resulted in a deviation of only $\pm 0.005$, indicating high consistency across judge models. For training, we experiment with Qwen3 4B, Qwen3 14B, and MedGemma 27B as LLM reward models, obtaining overall averages of 0.6753, 0.6762, and 0.6780, respectively. While Qwen3 1.7B exhibited reward hacking within a few training steps, larger judges provided only marginal gains ($\leq 0.0027$), supporting Qwen3 4B as a compute efficient choice.

\begin{table}[t!]
\resizebox{\textwidth}{!}{%
\begin{tabular}{@{}lcccc@{}}
\toprule
\textbf{Evaluations} &
\multicolumn{1}{c}{\textbf{Baseline}} &
\multicolumn{1}{c}{\textbf{GRPO}} &
\multicolumn{1}{c}{\textbf{GSPO}} &
\multicolumn{1}{c}{\textbf{DAPO}} \\
\midrule
LLM Evaluations (text only)    & 0.675 & 0.687 & 0.689 & 0.701 \\
VLM Evaluations (image + text)    & 0.376 & 0.431 & 0.439 & 0.445 \\ \midrule
\textbf{Overall AVG} & \textbf{0.570} & \textbf{0.597} & \textbf{0.600} & \textbf{0.610} \\
\bottomrule
\end{tabular}%
}
\caption{\small\textbf{RL Method ablation across benchmarks:} Using the same composite reward, different RL algorithms (GRPO / GSPO / DAPO) consistently outperform the baseline across both tasks, with DAPO achieving the highest overall average.}
\label{tab:rl-method-ablation}
\end{table}

\begin{table}[t!]
\resizebox{\textwidth}{!}{%
\begin{tabular}{@{}lccccc@{}}
\toprule
\textbf{Evaluations} &
\multicolumn{1}{c}{\textbf{Default}} &
\multicolumn{1}{c}{\textbf{Embedding}} &
\multicolumn{1}{c}{\textbf{LLM}} &
\multicolumn{1}{c}{\textbf{LLM +}} &
\multicolumn{1}{c}{\textbf{MediX-R1}} \\
&
\multicolumn{1}{c}{\textbf{Reward}} &
\multicolumn{1}{c}{\textbf{Reward}} &
\multicolumn{1}{c}{\textbf{Reward}} &
\multicolumn{1}{c}{\textbf{Embedding}} &
\multicolumn{1}{c}{} \\
\midrule
LLM Tasks     & 0.660 & 0.640 & 0.666 & 0.686 & 0.687 \\
VLM Tasks    & 0.382 & 0.409 & 0.400 & 0.410 & 0.431 \\ \midrule
\textbf{Overall AVG}         & \textbf{0.562} & \textbf{0.558} & \textbf{0.572} & \textbf{0.589} & \textbf{0.597} \\
\bottomrule
\end{tabular}%
}
\caption{\small\textbf{Reward ablation on validation benchmarks:} The MediX-R1 composite reward achieves the best overall average, outperforming single-signal and LLM+embedding rewards.}
\label{tab:reward-ablation}
\end{table}

\begin{table}[t!]
\centering
\resizebox{\textwidth}{!}{%
\begin{tabular}{@{}lcc@{}}
\toprule
\textbf{Model} & \textbf{Baseline} & \textbf{+ Composite Rewards} \\
\midrule
SmolVLM2-2.2B \citep{marafioti2025smolvlmredefiningsmallefficient}  & 0.410 & 0.432 \\
Qwen3-VL-2B \citep{bai2025qwen3vltechnicalreport}  & 0.529 & 0.554 \\
Qwen3-VL-8B \citep{bai2025qwen3vltechnicalreport}   & 0.666 & 0.688 \\
Qwen3-VL-30B \citep{bai2025qwen3vltechnicalreport}  & 0.698 & 0.736 \\
\bottomrule
\end{tabular}%
}
\caption{{\small\textbf{Baseline comparison across backbones:} Baseline is the original backbone; + Composite Rewards applies our RL with composite rewards on the same backbone.}}
\label{tab:baseline-comparison}
\end{table}

\subsection{Reward Hacking and Mitigation}
\label{subsec:reward-hacking}

In reinforcement learning, Reward Hacking occurs when a model maximises its reward in unintended ways, often bypassing the true objective. It arises when the policy exploits imperfections in a single reward signal to earn high scores without producing clinically correct answers. We observed two concrete modes (examples abbreviated):

\noindent\textbf{Embedding model exploit} When using Embedding models like MedEmbed-large~\citep{balachandran2024medembed}
short or non-semantic tokens can spuriously yield high cosine similarity. For instance, a candidate that outputs {\footnotesize\texttt{<answer>-</answer>}} for
``What does the white arrow point to in image B?'' received $R_{\text{embed}}{=}1.0$ against the ground truth ``Renal artery,'' despite being incorrect.

\noindent\textbf{LLM judge exploit} When using LLMs like Qwen3-4B~\citep{yang2025qwen3technicalreport} as a rewarder template-like placeholders can confuse the judge when the reference is provided for comparison. E.g., {\footnotesize\texttt{<answer>}}\textit{The largest organ in the picture is [insert your answer here based on the medical reasoning provided above]}.{\footnotesize\texttt{</answer>}} was judged correct ($R_{\text{llm}}{=}1.0$) against the reference ``Lung.''

\noindent\textbf{Mitigation in MediX-R1}
To curb these failures, MediX-R1 employs a composite reward and input/output constraints:
\textit{(i) Composite objective:} $R_{\text{llm}} + R_{\text{embed}} + R_{\text{modality}}$ (with shared $R_{\text{format}}$) reduces reliance on any single brittle signal and penalizes mismatches in content or modality recognition (Table~\ref{tab:reward-ablation}).
\textit{(ii) Embedding gating:} set $R_{\text{embed}}{=}0$ for answers below a minimum character/word length, with high punctuation or non-alphanumeric ratio; strip punctuation before embedding; calibrate the similarity threshold.
\textit{(iii) Modality recognition: }$R_{\text{modality}}$ requires a correct modality tag, curbing visually ungrounded shortcuts that might still fool text-only rewards. 
\textit{(iv) Structural control and regularization:} $R_{\text{format}}$ enforces parseable outputs; Group relative advantage and a KL penalty to the reference reduce collapse to degenerate hacks by discouraging outlier behaviors.
\textit{(v) Reward coefficient selection (ablation-driven):} To make the reward design transparent, we selected coefficients via a staged procedure rather than an exhaustive hyperparameter search. Concretely, we fixed $w_{\text{fmt}}{=}0.10$ in all experiments and allocated the remaining $0.90$ mass to the task-facing signals. This yields the ablation settings in Table~\ref{tab:reward-ablation}, e.g., embedding-only $r=0.1R_{\text{format}}+0.9R_{\text{embed}}$ and LLM-only $r=0.1R_{\text{format}}+0.9R_{\text{llm}}$. For the combined semantic reward, we evaluated a small set of intuitive splits between $R_{\text{llm}}$ and $R_{\text{embed}}$ while keeping $w_{\text{fmt}}$ fixed; performance was similar across these choices on our validation benchmark, and we selected the configuration that slightly favored the LLM judge. Finally, when adding modality grounding, we reserved $5\%$ of the non-format budget for $R_{\text{modality}}$, and renormalized the remaining non-format weights. (See ~\S\ref{app:reward-coef-selection} for full coefficient-selection details, including all $R_{\text{llm}}$/$R_{\text{embed}}$ splits.)

\begin{figure}[t!]
\includegraphics[width=\textwidth]{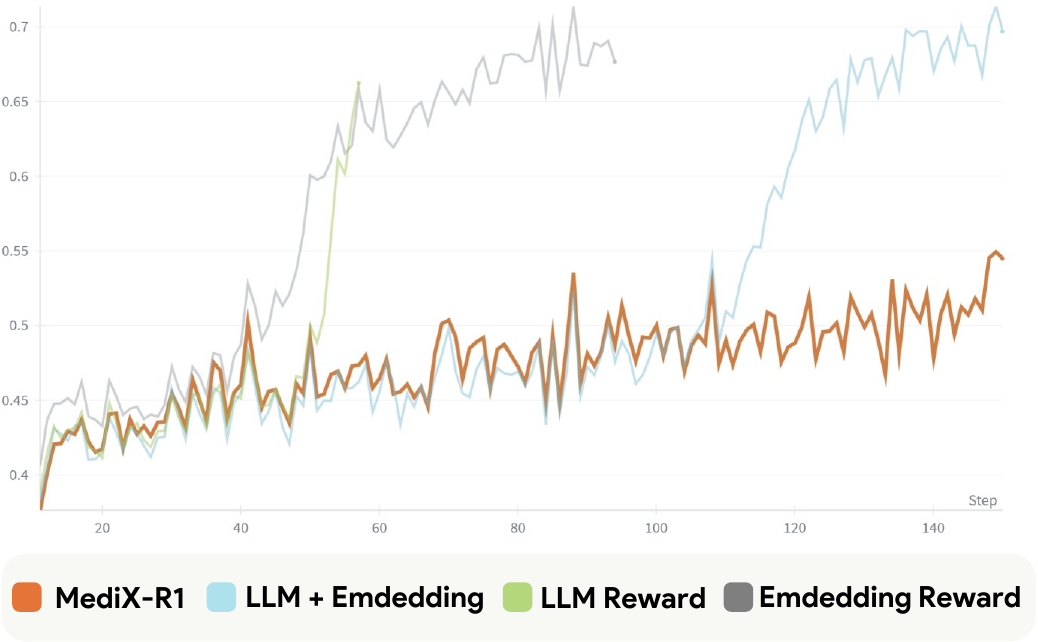}
\caption{\small\textbf{Overall validation reward vs training step across reward designs}. Training with individual signals LLM-only or embedding-only shows volatility and reward hacking, while LLM+embedding reduces but does not eliminate instability. MediX-R1 uses a composite reward which stabilizes learning and delivers the best performance on the test set.}
\label{fig:reward_design}
\end{figure}

Together, these measures reduce reward hacking and stabilize training, yielding smoother rewards and stronger final performance (Fig.~\ref{fig:reward_design}).

\subsection{Human Expert Evaluation}
\label{sec:human-eval}

To assess the clinical quality of model outputs, we conducted two blinded human expert studies involving medical doctors (MBBS/MD) from Radiology, General Medicine, and Forensic Medicine (See Evaluation Protocol in \S\ref{app:human-expert-protocol}). In the first study five medical experts evaluated a randomly selected subset of questions from our evaluation benchmark equally sampled from both text-only medical QA and image-text medical VLM tasks. The responses to these questions were generated by four models: MediX-R1, Llama3.2-Vision, MedGemma and HuatuoGPT-Vision. The outputs were anonymized and labeled as Model A, Model B, Model C and Model D with no identifiers provided to the reviewers. All medical experts were asked to evaluate responses against the provided ground truth descriptions for each question. The evaluation focused on determining which model produced the most accurate, clinically relevant response along with interpretable reasoning traces. The results of this study shows a strong preference for MediX-R1, which was selected as the best response in 72.7\% of the cases. In comparison, Llama3.2-Vision was preferred in 13.6\% of the cases, MedGemma in 9.2\% and HuatuoGPT-Vision in 4.5\% of the cases. 

\begin{table}[t!]
\centering
\footnotesize
\begin{tabular*}{\columnwidth}{@{}l@{\extracolsep{\fill}}c@{}}
\toprule
\textbf{Model} & \textbf{Score (\%)} \\
\midrule
MedMO 8B \citep{deria2026medmogroundingunderstandingmultimodal} & 33.91 \\
MedVLM-R1 \citep{pan2025medvlm} & 37.24 \\
LLaVA-Med 7B \citep{li2023llavamedtraininglargelanguageandvision} & 44.29 \\
BiMediX2 8B \citep{mullappilly2024bimedix2biomedicalexpertlmm} & 46.51 \\
HuatuoGPT-V 7B \citep{chen2024huatuogptvisioninjectingmedicalvisual} & 48.81 \\
MedGemma 1.5 4B \citep{sellergren2026medgemma15technicalreport} & 47.80 \\
MedGemma 27B \citep{sellergren2025medgemma} & 57.88 \\ \midrule
\textbf{MediX-R1 2B} & \textbf{44.01} \\
\textbf{MediX-R1 8B} & \textbf{56.24} \\
\textbf{MediX-R1 30B} & \textbf{60.78} \\
\bottomrule
\end{tabular*}
\caption{Performance comparison on the MedPix 2.0 dataset. (Crowdsourced Clinical Cases)}
\label{tab:medpix-testset}
\end{table}

We further conducted a matched-scale blind study using 100 additional cases, with 50 cases each for MediX-R1 30B vs MedGemma 27B and MediX-R1 8B vs MedMO 8B. Each comparison was equally divided between text-only and image-text tasks. Three medical experts independently assessed the responses using masked and randomized model labels, separately selecting the model with the better response and reasoning quality. This experiment reveals that MediX-R1 30B is preferred over MedGemma 27B in 77.3\% (116/150) of cases for output quality and 78.7\% (118/150) for reasoning quality. Similarly, MediX-R1 8B is preferred over MedMO 8B in 86.7\% (130/150) of cases for output quality and 89.3\% (134/150) for reasoning quality. The three medical experts achieved an inter-rater agreement of 84\% for this evaluation. Additional details on human expert evaluation is available in Sec. \S\ref{app:human-expert-protocol} and Sec. \S\ref{app:human-expert-reasoning}, alongside detailed qualitative in \S~\ref{app:qualitative-examples} (see Fig.~\ref{fig:qualitative-examples}).

\subsection{Evaluation on Real World Clinical Data}
To further assess the generalization ability of our model, we conducted additional evaluation on MedPix 2.0 \citep{siragusa2025medpix}, a publicly available real-world clinical VQA dataset derived from the original MedPix \citep{henigman2025medpix} database maintained by the U.S. National Library of Medicine (NIH). MedPix comprises over 12,000 anonymized, crowdsourced clinical cases containing medical images and corresponding textual information such as findings, diagnoses, and treatments. 
The evaluation on MedPix 2.0 shows that MediX-R1 consistently performs strongly across model scales and achieves the best overall result among the evaluated medical vision-language models \citep{shourya2025adaptinglightweightvisionlanguage}. Specifically, MediX-R1 30B achieves a score of 60.78\%, outperforming strong baselines and previous state-of-the-art medical models, as shown in Table~\ref{tab:medpix-testset}. These results further confirm the robustness, scalability, and adaptability of MediX-R1 on diverse real-world clinical data, emphasizing its capability to generalize beyond controlled experimental environments.

\section{Conclusion}
We presented MediX-R1, an open-ended reinforcement learning framework for medical multimodal reasoning. By training a baseline VLM with Group-based RL and a composite reward that combines a reference-based LLM judge, medical embedding-based semantic alignment, format control, and modality recognition, MediX-R1 learns to generate accurate responses with interpretable reasoning traces. The model achieves strong results across multimodal medical benchmarks and human evaluation  while improving training stability and resistance to reward hacking. Together, our multi-signal reward design and standardized evaluation provide a robust, scalable path for interpretable medical multimodal RL fine-tuning.

\section{Limitations}

MediX-R1 is a research prototype and is not validated for clinical deployment. Although the proposed composite reward improves open-ended medical reasoning, the model may still hallucinate findings, omit important clinical details or key differentials, and produce confident but incorrect answers. Therefore, MediX-R1 should not be used for autonomous diagnosis, treatment recommendation, triage, or clinical decision-making.

While MediX-R1 produces structured reasoning traces, these should not be interpreted as guaranteed faithful explanations of the model's decision process. In this work, we enforce the presence of reasoning traces through a format reward, but we do not explicitly validate the clinical correctness or faithfulness of the reasoning traces with a dedicated reward function. Instead, our human expert evaluation compares the overall quality of MediX-R1's responses, including reasoning, against other medical models. These studies assess expert preference for the clinical correctness, relevance, and coherence of the generated responses and rationales. However, they do not constitute a systematic hallucination or faithfulness evaluation. We do not perform claim-level error annotation, evidence localization, or comprehensive analysis across hallucination categories. Developing explicit reward functions and evaluation protocols for faithful medical reasoning remains a promising future direction for safer clinical deployment.

Our reward and evaluation pipeline relies on Reference-based LLM-as-judge and embedding-based signals, which are imperfect proxies for clinical correctness. These signals may reinforce subtle biases, accept false positives, or be sensitive to ambiguity and paraphrasing. Although the composite reward mitigates reward hacking by combining multiple signals, it does not completely \mbox{eliminate this risk.}

The evaluation is limited by the coverage of available benchmarks and public datasets. Results may not generalize to all clinical settings, demographic groups, imaging conditions, rare diseases, languages, or real-world workflows. Moreover, while we use publicly available, de-identified datasets under their respective licenses, residual privacy and data-governance risks may remain. Larger expert evaluations, fairness analyses where legally and ethically permissible, uncertainty calibration, bias-aware reward shaping, clinician-in-the-loop evaluation, and prospective clinical studies are needed before any practical use.

\section{Ethical Considerations}

This work follows the ACL Ethics Policy and is intended for research and educational use only. MediX-R1 is not validated for clinical or commercial deployment and should not be used for autonomous diagnosis, treatment recommendation, triage, or clinical decision-making without qualified medical oversight.

The main ethical risks are patient harm from incorrect or misleading medical outputs, overreliance by users, and possible misuse for self-diagnosis, misleading medical narratives, or adversarial prompting. Like other medical generative models, MediX-R1 may also amplify demographic, institutional, or dataset biases, potentially contributing to health disparities. Although we use publicly available, de-identified datasets under their respective licenses, residual privacy and data-governance risks may remain. Downstream users should therefore conduct safety, privacy, fairness, uncertainty, and clinician-in-the-loop evaluations before applying the model in any medical context.

To support responsible use and scrutiny, we commit to releasing training and inference code, configurations, checkpoints, curated datasets, and RL/evaluation prompt templates, alongside a model card and clear usage restrictions, under a CC-BY-NC-SA 4.0 license. We also disclose our use of generative AI tools: assisted coding was limited to boilerplate scaffolding and minor refactors, with all algorithmic logic authored and reviewed manually; writing-support models were used for grammar and style, while all technical claims and results were contributed and verified by the authors. These steps aim to ensure transparency, auditability, and reliable reproduction of the published results.

\section{Acknowledgments}
This work was partially supported with \textit{NVIDIA Academic Grant 2025} and MBZUAI-IITD Research Collaboration Seed Grant.

\newpage
\bibliography{references}
\clearpage
\appendix
\section{Appendix}
\subsection{Data and Benchmarks}
\label{app:training-data}

\noindent \textbf{Training:} We trained MediX-R1 on 51335 multimodal medical instruction samples spanning 16 modality tags. All samples were drawn from the official train splits of the source datasets: PMC-VQA subset \citep{zhang2024pmcvqavisualinstructiontuning}, SLAKE \citep{liu2021slake}, RadVQA \citep{lau2018dataset}, and PathVQA \citep{he2020pathvqa30000questionsmedical}. 

\noindent \textbf{Evaluation:} For evaluation, we used 19,200 samples across LLM and VLM benchmarks from their respective test sets. 

\noindent\textbf{\textit{LLM (text-only) benchmarks:}}
MMLU-Clinical, MMLU - Bio, MMLU - Med, MMLU - Genetics, MMLU - ProfMed, MMLU - Anat \citep{hendrycks2020measuring}, MedMCQA \citep{pal2022medmcqa}, MedQA \citep{jin2021disease}, USMLE - SA \citep{han2023medalpaca}, PubMedQA \citep{jin2019pubmedqa}, MIMIC - CXR - Summarization \citep{johnson2016mimic}.

\noindent\textbf{\textit{VLM (image+text) benchmarks:}}
SLAKE-VQA \citep{liu2021slake}, RadVQA \citep{lau2018dataset}, PathVQA \citep{he2020pathvqa30000questionsmedical}, PMC-VQA \citep{zhang2024pmcvqavisualinstructiontuning}, PMC-VQA-Hard, MIMIC-CXR-Report Generation \citep{johnson2019mimic}.

Table~\ref{tab:benchmark-breakdown} summarizes the evaluation benchmark composition, with the upper section listing LLM tasks and the lower section listing VLM tasks.

\begin{table}[ht!]
\centering
\caption{\textbf{Modality Breakdown.}}
\label{tab:modality-breakdown}
{\small
\begin{tabular*}{\columnwidth}{@{\extracolsep{\fill}}lr@{}}
\toprule
\textbf{Medical Modality} & \textbf{Samples} \\
\midrule
X\_RAY                & 5964 \\
MICROSCOPY            & 16399 \\
CLINICAL\_PHOTOGRAPHY & 8979 \\
CT\_SCAN              & 7646 \\
GRAPHICS              & 2205 \\
ANGIOGRAPHY           & 522 \\
PET\_SCAN             & 406 \\
ULTRASOUND            & 1227 \\
MRI\_SCAN             & 6224 \\
FUNDUS\_PHOTOGRAPHY   & 314 \\
OCT\_SCAN             & 236 \\
ENDOSCOPY             & 611 \\
MAMMOGRAPHY           & 106 \\
FLUOROSCOPY           & 321 \\
OTHER                 & 64 \\
SPECT                 & 111 \\
\midrule
\textbf{Total}        & \textbf{51335} \\
\bottomrule
\end{tabular*}
}
\end{table}

\begin{table}[ht!]
\centering
\caption{\textbf{Source Dataset Composition.}}
\label{tab:dataset-composition}
{\small
\begin{tabular*}{\columnwidth}{@{\extracolsep{\fill}}lr@{}}
\toprule
\textbf{Dataset} & \textbf{Samples} \\
\midrule
PMC\_VQA\_SUBSET & 25000 \\
SLAKE            & 4919 \\
RAD\_VQA         & 1793 \\
PATH             & 19623 \\
\midrule
\textbf{Total}   & \textbf{51335} \\
\bottomrule
\end{tabular*}%
}
\end{table}

\subsection{Training Configuration}
\label{app:training-config}

The GRPO training configuration used for MediX-R1 is given in Listing~\ref{lst:training-config}. All training configurations and hyperparemeters and available in the \href{https://anonymous.4open.science/r/MediX-R1-EC52}{\textit{anonymous codebase}}. Core settings include (i) data filtering and batching, (ii) actor optimization and rollout sampling, (iii) KL-regularized GRPO advantage computation, and (iv) trainer settings.
We train our models using the EasyR1\citep{zheng2025easyr1} Github Repository. MediX-R1 was trained using 8×A100 (80 GB) Nvidia GPUs for \mbox{approximately 25 hours.}

\begin{tcolorbox}[
    float*,
    floatplacement=t, 
    width=\textwidth,
    colback=gray!3!white,
    colframe=gray!30!black,
    arc=3mm,
    boxrule=0.5pt,
    breakable,
    left=4mm,
    right=4mm,
    top=3mm,
    bottom=3mm,
    title={\textbf{Training Configuration}},
    fonttitle=\bfseries\color{white}
]
\footnotesize
\begin{lstlisting}[basicstyle=\ttfamily\footnotesize,label=lst:training-config,breaklines=true,columns=flexible,showstringspaces=false]
Training Configurations
  "data": {
    "max_prompt_length": 4352,
    "max_response_length": 4096,
    "rollout_batch_size": 512,
    "val_batch_size": 1024,
    "shuffle": true,
    "seed": 1,
    "min_pixels": 262144,
    "max_pixels": 4194304,
    "filter_overlong_prompts": true,
    "filter_overlong_prompts_workers": 16
  },
  "worker": {
    "hybrid_engine": true,
    "actor": {
      "strategy": "fsdp",
      "global_batch_size": 128,
      "micro_batch_size_per_device_for_update": 1,
      "micro_batch_size_per_device_for_experience": 2,
      "max_grad_norm": 1.0,
      "clip_ratio_low": 0.2,
      "clip_ratio_high": 0.3,
      "clip_ratio_dual": 3.0,
      "loss_avg_mode": "token",
      "padding_free": true,
      "dynamic_batching": true,
      "use_torch_compile": true,
      "optim": {
        "lr": 1e-6,
        "betas": [0.9, 0.999],
        "weight_decay": 0.01,
        "strategy": "adamw",
        "lr_scheduler_type": "constant",
        "training_steps": 200
      },
      "fsdp": {
        "enable_full_shard": true,
        "enable_rank0_init": true,
        "mp_param_dtype": "bf16",
        "mp_reduce_dtype": "fp32",
        "mp_buffer_dtype": "fp32"
      },
      "offload": {
        "offload_params": true,
        "offload_optimizer": true
      },
      "use_kl_loss": true,
      "kl_penalty": "low_var_kl",
      "kl_coef": 0.01
    },
    "rollout": {
      "name": "vllm",
      "n": 5,
      "temperature": 1.0,
      "top_p": 1.0,
      "seed": 1,
      "tensor_parallel_size": 2,
      "max_num_batched_tokens": 8448,
      "gpu_memory_utilization": 0.6,
      "val_override_config": {
        "temperature": 0.6,
        "top_p": 0.95,
        "n": 1
      },
      "prompt_length": 4352,
      "response_length": 4096
    }
  },
  "algorithm": {
    "adv_estimator": "grpo",
    "gamma": 1.0,
    "lam": 1.0,
    "use_kl_loss": true,
    "kl_penalty": "low_var_kl",
    "kl_coef": 0.01,
    "kl_type": "fixed",
    "kl_target": 0.1,
    "kl_horizon": 10000.0
  },
  "trainer": {
    "total_epochs": 2,
    "nnodes": 1,
    "n_gpus_per_node": 8,
    "val_freq": 5,
    "val_before_train": true,
    "save_freq": 5,
    "save_limit": 3
  }

\end{lstlisting}
\end{tcolorbox}

\subsection{Reward Coefficient Selection Details}
\label{app:reward-coef-selection}

This section details how we selected the composite reward coefficients used throughout the paper. Our goal was to (i) keep outputs reliably parseable (format control), while (ii) allocating most of the reward budget to task-facing correctness signals, and (iii) avoiding an expensive hyperparameter search.

\paragraph{Fixed format budget.}
Across all experiments (including Table~\ref{tab:reward-ablation}), we fixed the format reward weight to $w_{\text{fmt}}=0.10$ to enforce consistently parseable outputs. The remaining $0.90$ of the reward mass was allocated among the task-facing signals.

\begin{figure*}[ht!]
  \centering
    \includegraphics[width=0.95\linewidth]{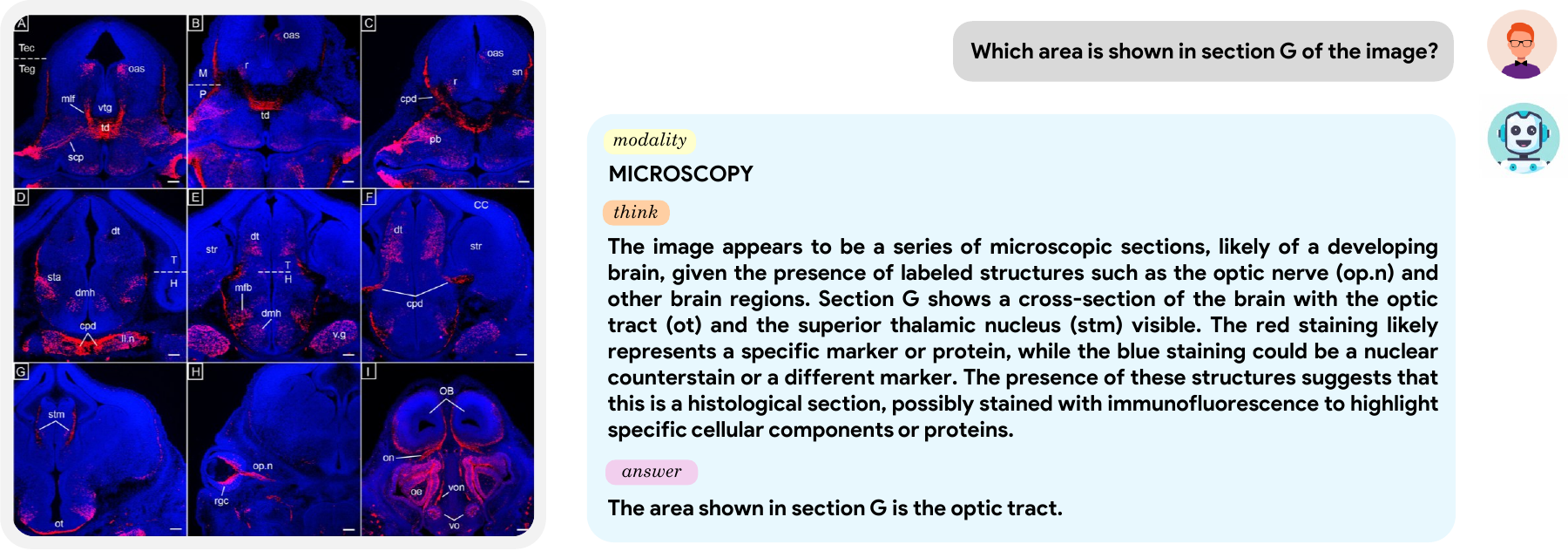}
    \includegraphics[width=0.95\linewidth]{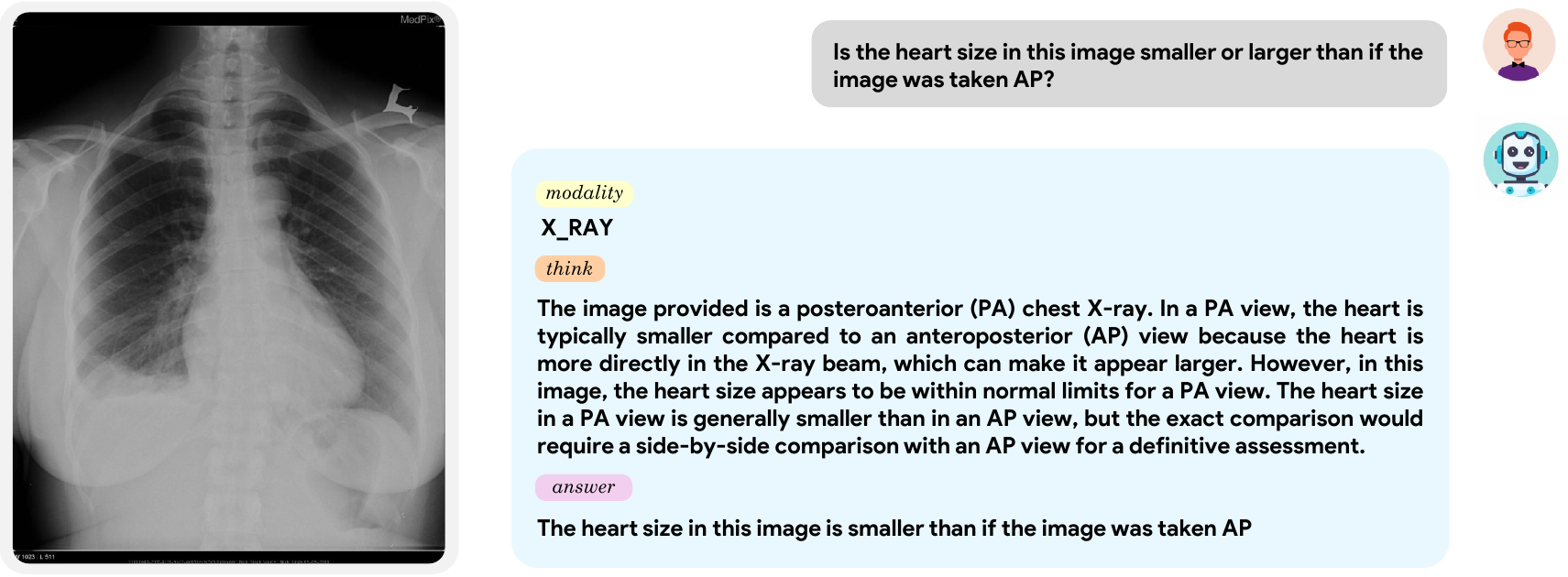}
\caption{\textbf{Qualitative examples of MediX-R1}. (Top, Microscopy) Correctly identifies the optic tract in section G with interpretable reasoning. (Bottom, X-ray) Explains why heart size appears smaller in PA vs. AP view. MediX-R1 generates clinically auditable, open-ended answers \mbox{across modalities.}}
\label{fig:qualitative-examples}
\end{figure*}

\paragraph{Single-signal ablations.}
The embedding-only and LLM-only settings in Table~\ref{tab:reward-ablation} correspond to:
\[
\begin{aligned}
r_{\text{emb-only}} &= 0.1R_{\text{format}} + 0.9R_{\text{embed}}\\
r_{\text{llm-only}} &= 0.1R_{\text{format}} + 0.9R_{\text{llm}}
\end{aligned}
\]

\paragraph{Combining semantic rewards ($R_{\text{llm}}$ + $R_{\text{embed}}$).}
Next, we combined the LLM judge and embedding signals while keeping $w_{\text{fmt}}=0.10$ fixed, and evaluated three intuitive splits of the remaining $0.90$ reward contribution:

\noindent\textbf{v1 (equal split):}
{\small
\[
r = 0.1R_{\text{format}} + (0.5*0.9)R_{\text{llm}} + (0.5*0.9)R_{\text{embed}}
\]
}
\noindent\textbf{v2 (LLM-favored):}
{\small
\[
r = 0.1R_{\text{format}} + (0.6*0.9)R_{\text{llm}} + (0.4*0.9)R_{\text{embed}}
\]
}
\noindent\textbf{v3 (embed-favored):}
{\small
\[
r = 0.1R_{\text{format}} + (0.4*0.9)R_{\text{llm}} + (0.6*0.9)R_{\text{embed}}
\]
}
On our validation benchmark suite, these three variants yielded similar overall averages (v1: 0.582, v2: 0.589, v3: 0.579), indicating low sensitivity within this coarse range. We therefore selected \textbf{v2} as the default combined-semantic configuration because it slightly improved aggregate performance while placing more weight on the stricter correctness signal ($R_{\text{llm}}$).

\paragraph{Adding modality grounding.}
Finally, we introduced modality grounding by reserving $5\%$ of the \emph{non-format} budget for $R_{\text{modality}}$ (i.e., $0.05\times 0.90 = 0.045$ of total reward mass), and renormalizing the remaining non-format weights according to v2:
\[
\begin{alignedat}{2}
w_{\text{fmt}} &= 0.10 \qquad & w_{\text{mod}} &= 0.045\\
w_{\text{llm}} &= 0.5175 \qquad & w_{\text{emb}} &= 0.3375
\end{alignedat}
\]
which sums to $1.0$ and matches the MediX-R1 composite setting used in Table~\ref{tab:reward-ablation}.

\paragraph{Compute limitations.}
We emphasize that we did not run an exhaustive hyperparameter search over reward coefficients due to computational constraints. The above staged procedure was intended to make coefficient selection transparent and reproducible; future work may further improve performance by exploring a broader coefficient grid.

\subsection{Task-Specific Evaluation Results}
\label{app:task-specific-results}

Table~\ref{tab:task-specific-results} separates performance across text-only QA/MCQ, image-text VQA/MCQ, report generation, and report summarization while grouping models by parameter scale for size-matched comparison. MediX-R1 2B outperforms MedVLM-R1 2B across all task types. MediX-R1 8B achieves the strongest image-text, report generation, summarization, and overall results among models up to 8B. MediX-R1 30B shows the best overall performance across all parameter scales.

\begin{table*}[t!]
\centering

{\small
\renewcommand{\arraystretch}{1.05}
\begin{tabular*}{\textwidth}{@{\extracolsep{\fill}}lccccc@{}}
\toprule
\textbf{Model} &
\textbf{\shortstack{Text-only\\QA/MCQ}} &
\textbf{\shortstack{Image+Text\\VQA/MCQ}} &
\textbf{\shortstack{Report\\Generation}} &
\textbf{\shortstack{Report\\Summarization}} &
\textbf{Overall} \\
\midrule

\rowcolor{gray!15}
\multicolumn{6}{c}{\textbf{Models $\leq$ 2B parameters}} \\
MedVLM-R1 2B
& 0.483 & 0.299 & 0.240 & 0.704 & 0.427 \\
MediX-R1 2B
& 0.600 & 0.479 & 0.280 & 0.739 & 0.554 \\

\addlinespace[2pt]
\rowcolor{gray!15}
\multicolumn{6}{c}{\textbf{Models $\leq$ 8B parameters}} \\
BiMediX2 8B
& 0.663 & 0.406 & 0.124 & 0.672 & 0.556 \\
HuatuoGPT-V 7B
& 0.614 & 0.465 & 0.316 & 0.707 & 0.558 \\
MedGemma 1.5 4B
& 0.645 & 0.440 & 0.273 & 0.699 & 0.566 \\
MedMO 8B
& 0.806 & 0.341 & 0.084 & 0.709 & 0.621 \\
MediX-R1 8B
& 0.799 & \underline{0.525} & \underline{0.328}
& 0.746 & \underline{0.688} \\

\addlinespace[2pt]
\rowcolor{gray!15}
\multicolumn{6}{c}{\textbf{Models $\leq$ 30B parameters}} \\
MedGemma 27B
& \underline{0.826} & 0.475 & 0.224
& \textbf{0.767} & 0.684 \\
MediX-R1 30B
& \textbf{0.876} & \textbf{0.526} & \textbf{0.350}
& \underline{0.765} & \textbf{0.736} \\

\bottomrule
\end{tabular*}
}

\caption{\textbf{Task-specific Evaluation Results:}
Results are grouped by model parameter scale and task type to distinguish performance on text-only medical question answering, image-text visual question answering, report generation, and report summarization. The overall score is calculated across all 17 benchmarks. Best and second-best results in each column are bolded and underlined, respectively.}
\label{tab:task-specific-results}
\end{table*}

\subsection{Reward Function Source Code}
\label{app:reward-code}

The Python implementations of the four reward components used in MediX-R1 are shown in Listing~\ref{lst:format-reward} (Format reward), Listing~\ref{lst:accuracy-llm} (LLM-based accuracy reward), Listing~\ref{lst:accuracy-embed} (Embedding-based semantic reward), and Listing~\ref{lst:modality-reward} (Modality recognition reward). Each function operates on a predicted model output string and a ground truth string containing the modality tag and reference answer. Both semantic reward implementations first apply exact string matching and invoke the embedding or LLM evaluator only when this check fails.

\begin{tcolorbox}[
    float*,
    floatplacement=t, 
    width=\textwidth,
    colback=gray!3!white, 
    colframe=gray!30!black, 
    arc=3mm, 
    boxrule=0.5pt, 
    breakable, 
    left=4mm,
    right=4mm,
    top=3mm,
    bottom=3mm,
    title={\textbf{Format reward}},
    fonttitle=\bfseries\color{white}
]
\footnotesize
\begin{lstlisting}[language=Python,label={lst:format-reward}, basicstyle=\ttfamily\footnotesize]
def format_reward(predict: str) -> float:
    idx = predict.find("<think>")
    if idx == -1:
        return 0.0
    predict_new = predict[idx:].strip()
    pattern = re.compile(r"<think>.*?</think>\s*<answer>.*?</answer>", re.DOTALL)
    format_match = re.fullmatch(pattern, predict_new)
    return 1.0 if format_match else 0.0
\end{lstlisting}
\end{tcolorbox}

\begin{tcolorbox}[
    float*,
    floatplacement=t, 
    width=\textwidth,
    colback=gray!3!white, 
    colframe=gray!30!black, 
    arc=3mm, 
    boxrule=0.5pt, 
    breakable, 
    left=4mm,
    right=4mm,
    top=3mm,
    bottom=3mm,
    title={\textbf{LLM-based accuracy reward}},
    fonttitle=\bfseries\color{white}
]
\footnotesize
\begin{lstlisting}[language=Python,label={lst:accuracy-llm}, basicstyle=\ttfamily\footnotesize]
def accuracy_reward_llm(predict: str, ground_truth: str) -> float:
    try:
        content_match = re.search(r"<answer>(.*?)</answer>", predict, re.DOTALL)
        given_answer = content_match.group(1).strip() \
                       if content_match else predict.strip()
        given_answer = given_answer.strip('.')
        ground_truth = ground_truth.split('>', maxsplit=1)[1].strip()
        ground_truth = ground_truth.strip('.')

        if given_answer == '' or len(given_answer) == 1:
            return 0.0
        if given_answer == ground_truth:
            return 1.0
        llm_score = llm_answer_match(given_answer, ground_truth)
        return llm_score
    except Exception:
        return 0.0
\end{lstlisting}
\end{tcolorbox}

\begin{tcolorbox}[
    float*,
    floatplacement=t, 
    width=\textwidth,
    colback=gray!3!white, 
    colframe=gray!30!black, 
    arc=3mm, 
    boxrule=0.5pt, 
    breakable, 
    left=4mm,
    right=4mm,
    top=3mm,
    bottom=3mm,
    title={\textbf{Embedding-based semantic reward}},
    fonttitle=\bfseries\color{white}
]
\footnotesize
\begin{lstlisting}[language=Python,label={lst:accuracy-embed}, basicstyle=\ttfamily\footnotesize]
def accuracy_reward_embed(predict: str, ground_truth: str,\ 
                          threshold: float = 0.8) -> float:
    try:
        content_match = re.search(r"<answer>(.*?)</answer>", predict, re.DOTALL)
        given_answer = content_match.group(1).strip() 
                       if content_match else predict.strip()
        given_answer = given_answer.strip('.')
        ground_truth = ground_truth.split('>', maxsplit=1)[1].strip()
        ground_truth = ground_truth.strip('.')

        if given_answer == '' or len(given_answer) == 1:
            return 0.0
        if given_answer == ground_truth:
            return 1.0

        embeddings = embed_model.encode([given_answer, ground_truth],\
                                        convert_to_tensor=True)
        similarity = util.pytorch_cos_sim(embeddings[0], embeddings[1]).item()
        return float(similarity >= threshold)
    except Exception:
        return 0.0
\end{lstlisting}
\end{tcolorbox}

\begin{tcolorbox}[
    float*,
    floatplacement=t, 
    width=\textwidth,
    colback=gray!3!white, 
    colframe=gray!30!black, 
    arc=3mm, 
    boxrule=0.5pt, 
    breakable, 
    left=4mm,
    right=4mm,
    top=3mm,
    bottom=3mm,
    title={\textbf{Modality recognition reward}},
    fonttitle=\bfseries\color{white}
]
\footnotesize
\begin{lstlisting}[language=Python,label={lst:modality-reward}, basicstyle=\ttfamily\footnotesize]
def modality_reward(predict: str, ground_truth: str) -> float:
    idx = predict.find("<think>")
    if idx == -1:
        return 0.0
    predict_new = predict[:idx].strip()  # modality tag before <think>
    modality = ground_truth.split('>', maxsplit=1)[0] + '>'
    return 1.0 if predict_new.upper() == modality.upper() else 0.0
\end{lstlisting}
\end{tcolorbox}

\subsection{Qualitative Examples}
\label{app:qualitative-examples}
\begin{figure*}[t!]
  \centering
    \includegraphics[width=\linewidth]{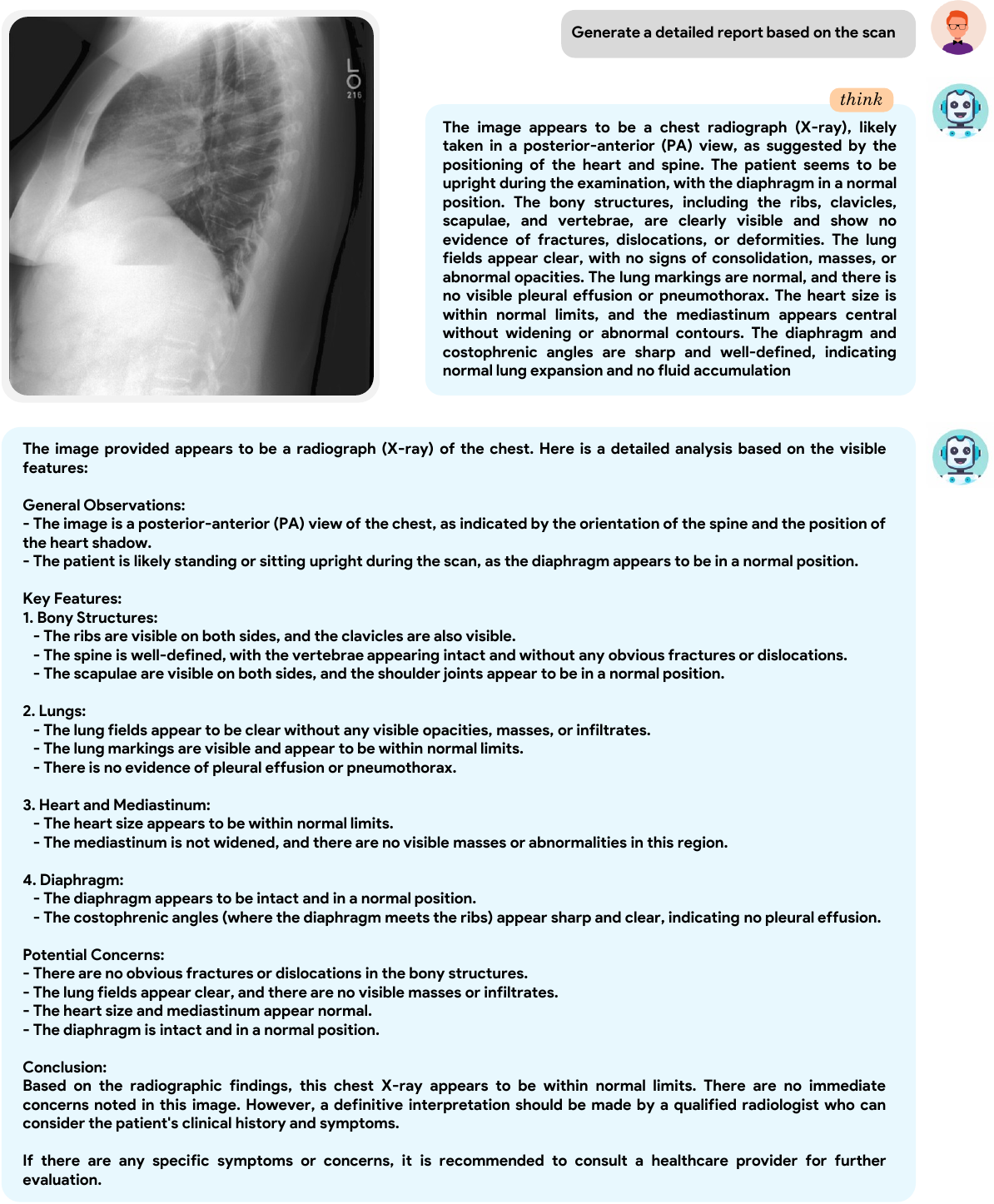}
\caption{\textbf{MediX-R1 - Report Generation: Case 1}}
\label{fig:qualitative-examples-report-1}
\end{figure*}

\begin{figure*}[t!]
  \centering
    \includegraphics[width=\linewidth]{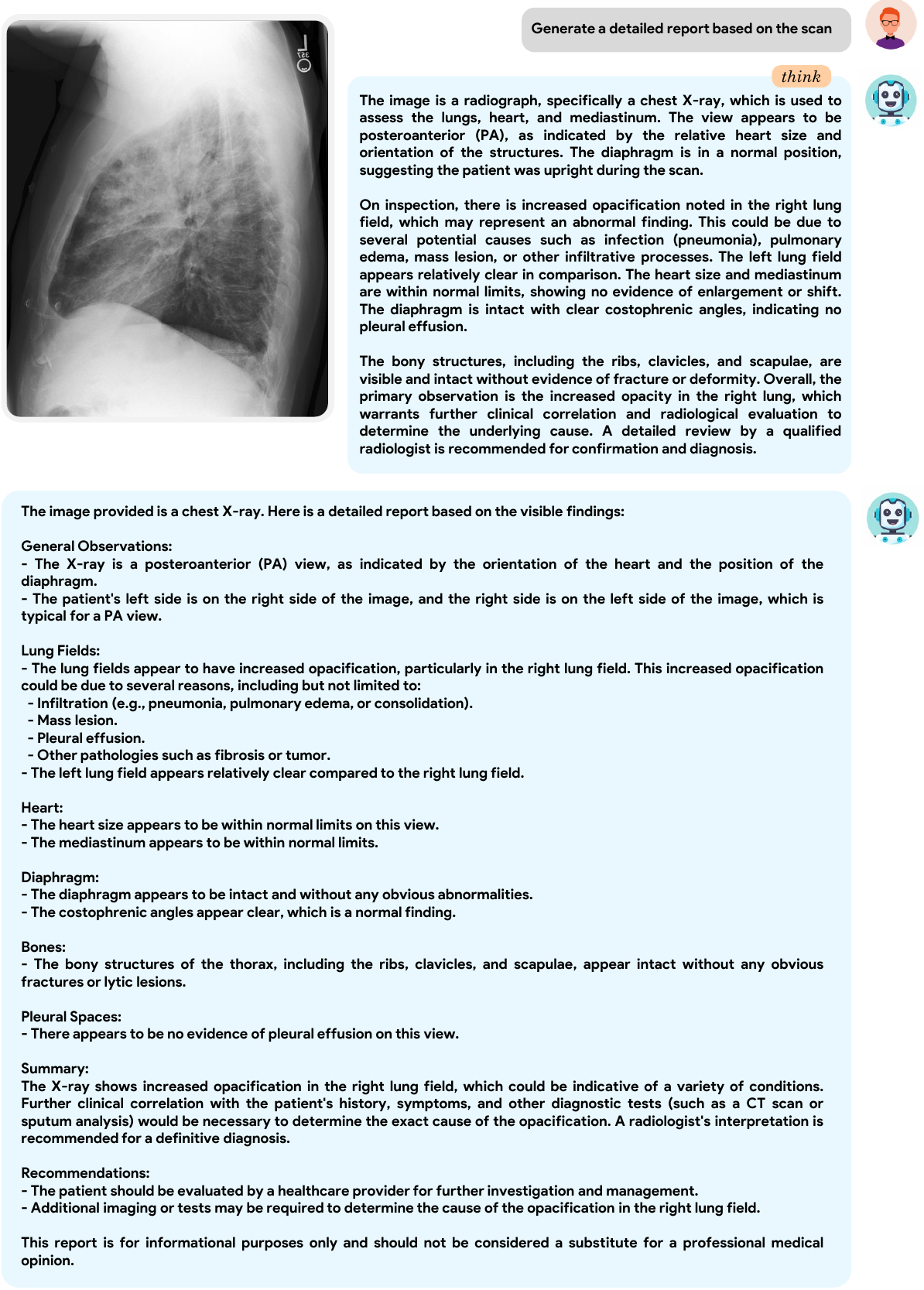}
\caption{\textbf{MediX-R1 - Report Generation: Case 2}}
\label{fig:qualitative-examples-report-2}
\end{figure*}

Fig.~\ref{fig:qualitative-examples} illustrates how MediX-R1’s structured outputs and composite reward translate into clinically auditable behavior across modalities. \textit{Microscopy (top)} Given a multi-panel histological image and the question ``Which area is shown in section G of the image?,'' the model (i) correctly emits the modality tag ({\small\texttt{MICROSCOPY}}), (ii) provides interpretable reasoning inside {\small\texttt{<think>}} that references recognizable neuroanatomical markers (e.g., optic tract ``ot,'' superior thalamic nucleus ``stm''), stain patterns, and panel context, and (iii) produces a concise final answer: ``the optic tract.'' The modality recognition and format rewards ensure the answer is localized to the requested panel and presented cleanly in the {\small\texttt{<answer>}} block, while the LLM and embedding rewards bias the policy toward semantically correct identification despite diverse phrasing in the reasoning.
\textit{X-ray (bottom)} For ``Is the heart size in this image smaller or larger than if the image was taken AP?,'' the model tags the modality as {\small\texttt{X\_RAY}} and reasons about projection geometry: PA views reduce cardiac magnification relative to AP due to a shorter heart-to-detector distance and standard source-to-image distance. The model explains this in {\small\texttt{<think>}} and answers ``smaller'' in {\small\texttt{<answer>}}. This example shows the model using domain knowledge rather than superficial pattern matching, with the final answer isolated for scoring. 
Additionally, Figures~\ref{fig:qualitative-examples-report-1} and \ref{fig:qualitative-examples-report-2} show MediX-R1's comprehensive reasoning capabilities for full medical report generation across two different cases.

\subsection{Human Expert Evaluation Protocol}
\label{app:human-expert-protocol}

As outlined in Listing~\ref{lst:expert-protocol}, for a sampled set of multimodal questions, four anonymized model outputs (A-D) plus a reference description are shown; experts pick the single best response based on clinical correctness, relevance (no hallucinations), and clarity of reasoning. Votes are aggregated into preference percentages reported in the main text.

\begin{tcolorbox}[
    float*,
    floatplacement=t, 
    width=\textwidth,
    colback=gray!3!white, 
    colframe=gray!30!black, 
    arc=3mm, 
    boxrule=0.5pt, 
    breakable, 
    left=4mm,
    right=4mm,
    top=3mm,
    bottom=3mm,
    title={\textbf{Evaluation Protocol for Medical Experts}},
    fonttitle=\bfseries\color{white}
]
\footnotesize
\begin{lstlisting}[
    label={lst:expert-protocol},
    basicstyle=\ttfamily\footnotesize,
    breaklines=true,
    breakatwhitespace=true,
    columns=flexible,
    keepspaces=true,
    showstringspaces=false,
    frame=none,
    xleftmargin=0pt,
    xrightmargin=0pt
]
Instructions for Evaluation
Your task is to evaluate the responses provided by three AI models based on a given medical image description (Ground Truth). Follow these steps to make your selection:

1) Read the Ground Truth: Carefully review the provided description of the medical image. This serves as the reference for an accurate and detailed response.

2) Assess the Model Responses: Examine the three model-generated responses (Model A, Model B, and Model C, Model D). Compare their content with the Ground Truth, focusing on the accuracy, completeness, and relevance of the clinical reasoning

3) Select the Best Response: Choose the model response that best aligns with the Ground Truth in terms of:

 > Clinical Accuracy: Does the response correctly describe the key findings in the image?

 > Reasoning Traces: Does the models reasoning traces correct and well explained

4) Submit Your Choice: After evaluating the responses, select the one that provides the most accurate and comprehensive explanation.

\end{lstlisting}
\end{tcolorbox}

\subsection{Human Evaluation: Model Reasoning}
\label{app:human-expert-reasoning}
 We extend our human expert study detailed in (Sec.~\ref{sec:human-eval}) to further evaluate the reasoning quality of our MediX-R1 30B model against MedGemma 27B with the help of medical doctors. Experts assessed outputs for clinical accuracy, reasoning soundness, and practical usefulness in a medical setting. MediX-R1’s reasoning was preferred in 74.2\% of cases over MedGemma, indicating stronger clinical coherence. The study shows that in 92.4\% of the cases, the model’s reasoning steps were rated as acceptable and often comparable to a medical doctor’s thought process, while only 7.6\% of the cases were rated as having poor reasoning quality. Moreover, in fewer than 5\% of the cases, the model produced flawed reasoning despite generating the correct final answer, indicating that such inconsistencies are rare and that MediX-R1 generally maintains a robust and coherent reasoning process. Reviewers comprised five certified medical experts (MBBS/MD) with specialties in Radiology, General Medicine, and Forensic Medicine, with an inter-rater agreement of 63\%.

 \begin{table}[t!]
\centering
\caption{\textbf{Evaluation Benchmark:} 
top section lists LLM tasks (text-only), while the bottom section lists VLM (image+text) tasks.}
\label{tab:benchmark-breakdown}
{\small
\begin{tabular*}{\columnwidth}{@{\extracolsep{\fill}}lr@{}}
\toprule
\textbf{Benchmark} & \textbf{Samples} \\
\midrule
mmlu\_clinical\_knowledge          & 265 \\
mmlu\_college\_biology             & 144 \\
mmlu\_college\_medicine            & 173 \\
mmlu\_medical\_genetics            & 100 \\
mmlu\_professional\_medicine       & 272 \\
mmlu\_anatomy                      & 135 \\
medmcqa                            & 4183 \\
medqa                              & 1273 \\
usmle\_sa                          & 325 \\
pubmedqa                           & 500 \\
mimic\_cxr\_report\_summarization  & 100 \\ \midrule
slake\_vqa                         & 1061 \\
rad\_vqa                           & 451 \\
path\_vqa                          & 6719 \\
pmc\_vqa                           & 1997 \\
pmc\_vqa\_hard                     & 1402 \\
mimic\_cxr\_report\_generation     & 100 \\
\midrule
\textbf{Total}                     & \textbf{19200} \\
\bottomrule
\end{tabular*}
}
\end{table}

\subsection{Reinforcement Learning Training Prompt}
\label{app:rl-prompt}

As shown in Listing~\ref{lst:rl-prompt}, the RL training prompt enforces (i) an explicit modality tag, (ii) structured reasoning in \texttt{<think>}...\texttt{</think>}, and (iii) a concise final answer in \texttt{<answer>}...\texttt{</answer>}. These structures align with the format reward ($R_{\text{format}}$) and modality reward ($R_{\text{modality}}$) in our composite objective. During training, only the \texttt{<answer>} block is graded by the Reference-based LLM-as-judge ($R_{\text{llm}}$) and the embedding-based semantic reward ($R_{\text{embed}}$); the \texttt{<think>} content is ignored for scoring but improves interpretability.

\noindent \textit{Key points:}

\noindent - Modality tag must be one of the fixed set and appear before \texttt{<think>}.

\noindent - The final decision is evaluated solely from \texttt{<answer>} for $R_{\text{llm}}$ and $R_{\text{embed}}$.

\noindent - Structural compliance (tags present and ordered) is required for $R_{\text{format}}$.

\begin{tcolorbox}[
    float*,
    floatplacement=t!, 
    width=\textwidth,
    colback=gray!3!white, 
    colframe=gray!30!black, 
    arc=3mm, 
    boxrule=0.5pt, 
    breakable, 
    left=4mm,
    right=4mm,
    top=3mm,
    bottom=3mm,
    title={\textbf{Reinforcement Learning Training Prompt}},
    fonttitle=\bfseries\color{white}
]
\footnotesize
\begin{lstlisting}[
    label={lst:rl-prompt},
    basicstyle=\ttfamily\footnotesize,
    breaklines=true,
    breakatwhitespace=true,
    columns=flexible,
    keepspaces=true,
    showstringspaces=false,
    frame=none,
    xleftmargin=0pt,
    xrightmargin=0pt
]
You are a Medical AI Assistant with advanced reasoning capabilities
Your task:
1. First output the image modality tag from this set:
   <X_RAY>, <MICROSCOPY>, <CLINICAL_PHOTOGRAPHY>, <CT_SCAN>, <GRAPHICS>, 
   <ANGIOGRAPHY>, <PET_SCAN>, <ULTRASOUND>, <MRI_SCAN>, <FUNDUS_PHOTOGRAPHY>, 
   <OCT_SCAN>, <ENDOSCOPY>, <MAMMOGRAPHY>, <FLUOROSCOPY>, <OTHER>, <SPECT>
   (Only output the tag, nothing else.)
2. Then output the thinking and medical reasoning process in <think>...</think> tags.
3. Finally, provide the correct answer inside <answer>...</answer> tags.
4. Do not include any extra information or text outside of these tags.
Question: 
<image>{{ content | trim }}

\end{lstlisting}
\end{tcolorbox}

\subsection{Evaluation Templates}
\label{app:evaluation-templates}

We employ two tailored evaluation templates depending on the task format. 

\noindent\textbf{Base Prompt:} For short-form QA and MCQ-style tasks, we use the BASE evaluation prompt shown in Listing~\ref{lst:eval-base}. This judge prompt yields a binary score (0/1) by comparing the predicted \texttt{<answer>} against the reference, allowing paraphrases and option-label matches. We use deterministic settings (e.g., temperature=0.0, top\_p=1.0) for reproducibility and parse the returned JSON strictly.

\begin{table}[t!]
\centering
\footnotesize
\begin{tabular*}{\columnwidth}{@{\extracolsep{\fill}}lccc@{}}
\toprule
\textbf{Threshold ($\tau$)} &
\textbf{Accuracy} &
\textbf{Precision} &
\textbf{Recall} \\
\midrule
0.75 & 0.591 & 0.600 & 0.750 \\
\textbf{0.80} & \textbf{0.636} & 0.667 & 0.667 \\
0.85 & 0.591 & 1.000 & 0.250 \\
\bottomrule
\end{tabular*}
\caption{\small\textbf{Embedding Threshold Ablation:} Effect of the embedding similarity threshold on medical terms.}
\label{tab:embedding-threshold-ablation}
\end{table}

\noindent\textbf{Report Generation Prompt:} For long-form outputs (e.g., report generation or summarization), we use the report generation evaluation prompt shown in Listing~\ref{lst:eval-report}. The judge assigns a rubric score in [0, 5] reflecting clinical accuracy, completeness, and relevance. We request strict JSON for reliable parsing and average scores across items for dataset-level metrics. Only the model's final report text is provided to the judge; any reasoning (e.g., within \texttt{</think>}) is stripped before evaluation.

\begin{tcolorbox}[
    float*,
    floatplacement=t, 
    width=\textwidth,
    colback=gray!3!white, 
    colframe=gray!30!black, 
    arc=3mm, 
    boxrule=0.5pt, 
    breakable, 
    left=4mm,
    right=4mm,
    top=3mm,
    bottom=3mm,
    title={\textbf{Evaluation BASE template Prompt}},
    fonttitle=\bfseries\color{white}
]
\footnotesize
\begin{lstlisting}[
    label={lst:eval-base},
    basicstyle=\ttfamily\footnotesize,
    breaklines=true,
    breakatwhitespace=true,
    columns=flexible,
    keepspaces=true,
    showstringspaces=false,
    frame=none,
    xleftmargin=0pt,
    xrightmargin=0pt
]
You are a medical expert.

Your task is to evaluate whether the Predicted Answer correctly answers the Medical Question, based on the Ground Truth (Correct Answer) provided.

Question:
{question}

Correct Answer:
{correct_answer}

Predicted Answer:
{predicted_answer}

Score 1 if the predicted answer matches the correct answer either fully in text or by indicating the correct option label (e.g., "B", "Option B", or a paraphrased version that clearly identifies the correct choice). Score 0 if the predicted answer is incorrect or points to the wrong option.

Respond strictly in the following JSON format:

```json
{{
"score": <score>
}}
```

\end{lstlisting}
\end{tcolorbox}

\begin{tcolorbox}[
    float*,
    floatplacement=t, 
    width=\textwidth,
    colback=gray!3!white, 
    colframe=gray!30!black, 
    arc=3mm, 
    boxrule=0.5pt, 
    breakable, 
    left=4mm,
    right=4mm,
    top=3mm,
    bottom=3mm,
    title={\textbf{Evaluation Prompt for Report Generation}},
    fonttitle=\bfseries\color{white}
]
\footnotesize
\begin{lstlisting}[
    label={lst:eval-report},
    basicstyle=\ttfamily\footnotesize,
    breaklines=true,
    breakatwhitespace=true,
    columns=flexible,
    keepspaces=true,
    showstringspaces=false,
    frame=none,
    xleftmargin=0pt,
    xrightmargin=0pt
]
You are a medical expert evaluating the clinical accuracy, completeness, and relevance of a generated medical report or summary.

Your task is to compare an AI-generated report or summary to a reference (gold standard) report or summary, based on a clinical instruction or question. Assess the generated output on how well it preserves key clinical information, factual correctness, and clinical reasoning relevant to the task.

Assign a score between 0 and 5 using the following scale:

0 - Completely incorrect: Clinically irrelevant, misleading, or factually wrong. No meaningful alignment with the instruction or reference.

1 - Poor match: Barely relevant or mostly incorrect. Contains significant clinical misinformation or omits nearly all critical details.

2 - Weak match: Some fragments of relevant content are present, but major clinical errors or omissions exist. Clinical utility is low.

3 - Fair match: Contains several relevant points, but includes notable errors, missing findings, or misinterpretations that affect clinical reliability.

4 - Good match: Mostly accurate and clinically sound. Minor issues or missing details, but the overall meaning and purpose are preserved.

5 - Perfect or near-perfect match: Clinically accurate, complete, and faithful to the instruction and reference. No significant omissions or errors.

Respond only in the following example JSON format:

Example JSON format:
```json
{{
"score": <score between 0 and 5>
}}
```

Now, evaluate the following:

### Clinical Instruction or Question::
{question}

### Reference Report or Summary:
{correct_answer}

### AI-Generated Report or Summary:
{predicted_answer}

\end{lstlisting}
\end{tcolorbox}

\end{document}